\documentclass[11pt,a4paper]{article}
\usepackage[acceptedWithA]{tacl2018v2}
\usepackage{algorithm}
\usepackage{algorithmicx}
\usepackage{amsmath}
\usepackage{booktabs}
\usepackage{caption}
\usepackage{hyperref}
\usepackage{graphicx}
\usepackage{multirow}
\usepackage{multicol}
\usepackage{tabularx}
\usepackage{times,latexsym}
\usepackage{url}
\usepackage[T1]{fontenc}

\usepackage{etoolbox}
\makeatletter
\patchcmd\@combinedblfloats{\box\@outputbox}{\unvbox\@outputbox}{}{%
   \errmessage{\noexpand\@combinedblfloats could not be patched}%
}%
 \makeatother

\usepackage{xspace,mfirstuc,tabulary}

\newcommand{\biaffine}{\textsc{b}i\textsc{affine}}

\newif\iftaclinstructions
\taclinstructionsfalse 
\iftaclinstructions
\renewcommand{\confidential}{}
\renewcommand{\anonsubtext}{(No author info supplied here, for consistency with
TACL-submission anonymization requirements)}
\newcommand{\instr}
\fi

\iftaclpubformat 

\else

\fi

\newcommand{\com}[1]{}

\title{Deep Contextualized Self-training for Low Resource Dependency Parsing}

\author{Guy Rotman \and Roi Reichart\\
Faculty of Industrial Engineering and Management, Technion, IIT\\
\texttt{grotman@campus.technion.ac.il} \\
\hspace{1.2em}\texttt{roiri@technion.ac.il} \\
}

\date{}

\begin{document}
\maketitle

\begin{abstract}

Neural dependency parsing has proven very effective, achieving state-of-the-art results on numerous domains and languages. Unfortunately, it requires large amounts of labeled data, that is costly and laborious to create. In this paper we propose a self-training algorithm that alleviates this annotation bottleneck by training a parser on its own output. Our \textit{Deep Contextualized Self-training (DCST)} algorithm utilizes representation models trained on sequence labeling tasks that are derived from the parser's output when applied to unlabeled data, and integrates these models with the base parser through a gating mechanism.  We conduct experiments across multiple languages, both in low resource in-domain and in cross-domain setups, and demonstrate that DCST substantially outperforms traditional self-training as well as recent semi-supervised training methods. \footnote{Our code is publicly available at \url{https://github.com/rotmanguy/DCST}.} \footnote{This paper was accepted to TACL in September 2019. }
\end{abstract}

\section{Introduction}

Deep Neural Networks (DNNs) have improved the state-of-the-art in a variety of NLP tasks. These include dependency parsing \cite{Dozat:17}, semantic parsing \cite{Hershcovich:17}, named entity recognition \cite{Yadav:18}, POS tagging \cite{plank-agic-2018-distant}, and machine translation \cite{vaswani:17}, among others. 

Unfortunately, DNNs rely on in-domain labeled training data, which is costly and laborious to achieve. This annotation bottleneck  limits the applicability of NLP technology to a small number of languages and domains. It is hence not a surprise that substantial recent research efforts have been devoted to DNN training based on both labeled and unlabeled data, which is typically widely available (\S~\ref{sec:previous}).   

A prominent technique for training machine learning models on labeled and unlabeled data is self-training \cite{yarowsky1995unsupervised,abney2004understanding}. In this technique, after the model is trained on a labeled example set it is applied to another set of unlabeled examples, and the automatically and manually labeled sets are then combined in order to re-train the model -- a process that is sometimes performed iteratively. While self-training has shown useful for a variety of NLP tasks, its success for deep learning models has been quite limited (\S~\ref{sec:previous}). 

Our goal is to develop a self-training algorithm that can substantially enhance DNN models in cases where labeled training data is scarce. Particularly, we are focusing (\S~\ref{sec:setups}) on the lightly supervised setup where only a small in-domain labeled dataset is available, and on the domain adaptation setup where the labeled dataset may be large but it comes from a different domain than the one to which the model is meant to be applied. Our focus task is dependency parsing, which is essential for many NLP tasks \citep{levy2014dependency,Angeli15,Toutanova16,Hadiwinoto17,Marcheggiani17}, but where self-training has typically failed (\S~\ref{sec:previous}). Moreover, neural dependency parsers \cite{Kiperwasser16, Dozat:17} substantially outperform their linear predecessors, which makes the development of self-training methods that can enhance these parsers in low-resource setups a crucial challenge.

We present a novel self-training  method, suitable for neural dependency parsing. Our algorithm (\S~\ref{sec:algorithm}) follows recent work that has demonstrated the power of pre-training for improving DNN models in NLP \cite{peters-etal-2018-deep,devlin2019bert} and particularly for domain adaptation \cite{ziser-reichart-2018-pivot}. However,  while in previous work a representation model, also known as a contextualized embedding model, is trained on a language modeling related task, our algorithm utilizes a representation model that is trained on sequence prediction tasks derived from the parser's output. 
Our representation model and the base parser are integrated into a new model through a gating mechanism, and the resulting parser is then trained on the manually labeled data.



We experiment (\S~\ref{sec:experiments},\ref{sec:results}) with a large variety of lightly-supervised and domain adaptation dependency parsing setups. For the lightly-supervised case we consider 17 setups: 7 in different English domains and 10 in other languages. For the domain adaptation case we consider 16 setups: 6 in different English domains and 10 in 5 other languages. Our Deep Contextualized Self-training (DCST) algorithm demonstrates substantial performance gains over a variety of baselines, including traditional self-training and the recent cross-view training approach (CVT) \cite{clark2018semi} that was designed for semi-supervised learning with DNNs.

\section{Previous Work}
\label{sec:previous}

\paragraph{Self-training in NLP} 

Self-training has shown useful for various NLP tasks, including word sense disambiguation \cite{mihalcea2004co, yarowsky1995unsupervised}, bilingual lexicon induction \cite{artetxe-etal-2018-robust}, neural machine translation \cite{imamura2018nict}, semantic parsing \cite{Goldwasser:11} and sentiment analysis \cite{he2011self}. For constituency parsing, self-training has shown to improve linear parsers both when large training data is available \cite{mcclosky2006effective,mcclosky2006reranking}, and in the lightly supervised and the cross-domain setups \cite{reichart2007self}. While several authors failed to demonstrate the efficacy of self-training for dependency parsing (e.g. \cite{rush2012improved}), recently it was found useful for neural dependency parsing in fully supervised multilingual settings \cite{rybak2018semi}.

The impact of self-training on DNNs is less researched compared to the extensive investigation with linear models. Recently, \newcite{ruder2018strong} evaluated the impact of self-training and the closely related tri-training method \cite{Zhou:05,Sogaard:10} on DNNs for part-of-speech (POS) tagging and sentiment analysis. They found self-training to be effective for the sentiment classification task, but it failed to improve their BiLSTM POS tagging architecture. Tri-training has shown effective for both the classification and the sequence tagging task, and in \newcite{Vinyals:15} it has shown useful for neural constituency parsing . This is in-line with \newcite{steedman2003bootstrapping} that demonstrated the effectiveness of the closely related co-training method \cite{Blum:98} for linear constituency parsers.

Lastly, \citet{clark2018semi} presented the Cross-view Training (CVT) algorithm, a variant of self-training that employs unsupervised representation  learning. CVT differs from classical self-training in the way it exploits the unlabeled data: it trains auxiliary models on restricted views
of the input to match the predictions of the full model that observes the whole input. 

We propose a self-training algorithm based on deep contextualized embeddings, where the embedding model is trained on sequence tagging tasks that are derived from the parser's output on unlabeled data. In extensive lightly supervised and cross-domain experiments with a neural dependency parser, we show that our DCST algorithm outperforms traditional self-training and CVT.


\paragraph{Pre-training and Deep Contextualized Embedding}

Our DCST algorithm is related to recent work on DNN pre-training. In this line, a DNN is first trained on large amounts of unlabeled data and is then used as the word embedding layer of a more complex model that is trained on labeled data to perform an NLP task. Typically, only the upper, task specific, layers of the final model are trained on the labeled data, while the parameters of the pre-trained embedding network are kept fixed.

The most common pre-training task is language modeling or a closely related variant \cite{mccann2017learned, peters-etal-2018-deep, devlin2019bert, ziser-reichart-2018-pivot}. The outputs of the pre-trained DNN are often referred to as contextualized word embeddings, as these DNNs typically generate a vector embedding for each input word, which takes its context into account. Pre-training has led to  performance gains in many NLP tasks.

Recently, \newcite{che2018towards} incorporated ELMo embeddings \cite{peters-etal-2018-deep} into a neural dependency parser and reported improvements over a range of Universal Dependency (UD) \cite{mcdonald2013universal,nivre2016universal, nivre2018universal} languages in the fully supervised setup. In this paper we focus on the lightly supervised and domain adaptation setups, trying to compensate for the lack of labeled data by exploiting automatically labeled trees generated by the base parser for unlabeled sentences. 

Our main experiments (\S \ref{sec:results}) are with models that utilize non-contextualized word embeddings. We believe this is a more practical setup when considering multiple languages and domains. Indeed, \newcite{che2018towards}, who trained their ELMo model on the unlabeled data of the CoNLL 2018 shared task, reported that "The training of ELMo on one language takes roughly 3 days on an NVIDIA P100 GPU.". However, we also demonstrate the power of our models when ELMo embeddings are available (\S \ref{sec:ablation}), in order to establish the added impact of deep contextualized self-training on top of contextualized word embeddings.

\com{We compare our DCST algorithm to a model that utilizes a deep contextualized embedding model trained on a language modeling task. However, to avoid the need to train a heavy contextualized embedding model for each language and domain, the baseline contextualized embedding model in our experiment is based on a BiLSTM - the same architecture we train on the automatically labeled trees in DCST. We believe this is a more practical setup when considering multiple languages and domains. Indeed, \newcite{che2018towards}, who trained their ELMo model on the unlabeled data of the CoNLL 2018 shared task, reported that "The training of ELMo on one language takes roughly 3 days on an NVIDIA P100 GPU.".}

\paragraph{Lightly Supervised Learning and Domain Adaptation for Dependency Parsing}

Finally, we briefly survey earlier attempts to learn parsers in setups where labeled data from the domain to which the parser is meant to be applied is scarce. We exclude from this brief survey literature that has already been mentioned above.

Some notable attempts are: exploiting short dependencies in the parser's output when applied to large target domain unlabeled data \cite{chen2008learning}, adding inter-sentence consistency constraints at test time \cite{rush2012improved}, selecting effective training domains \cite{plank2011effective}, exploiting parsers trained on different domains through a mixture of experts \cite{mcclosky2010automatic}, embedding features in a vector space \cite{chen2014feature}, and Bayesian averaging of a range of parser parameters \cite{shareghi2019bayesian}.

Recently, \citet{sato2017adversarial} presented an adversarial model for cross-domain dependency parsing in which the encoders of the source and the target domains are integrated through a gating mechanism. Their approach requires target domain labeled data for parser training and hence it cannot be applied in the unsupervised domain adaptation setup we explore (\S~\ref{sec:setups}). We adopt their gating mechanism to our model and extend it to integrate more than two encoders into a final model.

\section{Background: The \biaffine\ Parser}
\label{sec:biaffine}

The parser we utilize in our experiments is the \biaffine\ parser \cite{Dozat:17}. Since the  structure of the parser affects our DCST algorithm, we briefly describe it here.

\begin{figure}[!t]
\centering
\includegraphics[width=0.99\linewidth]{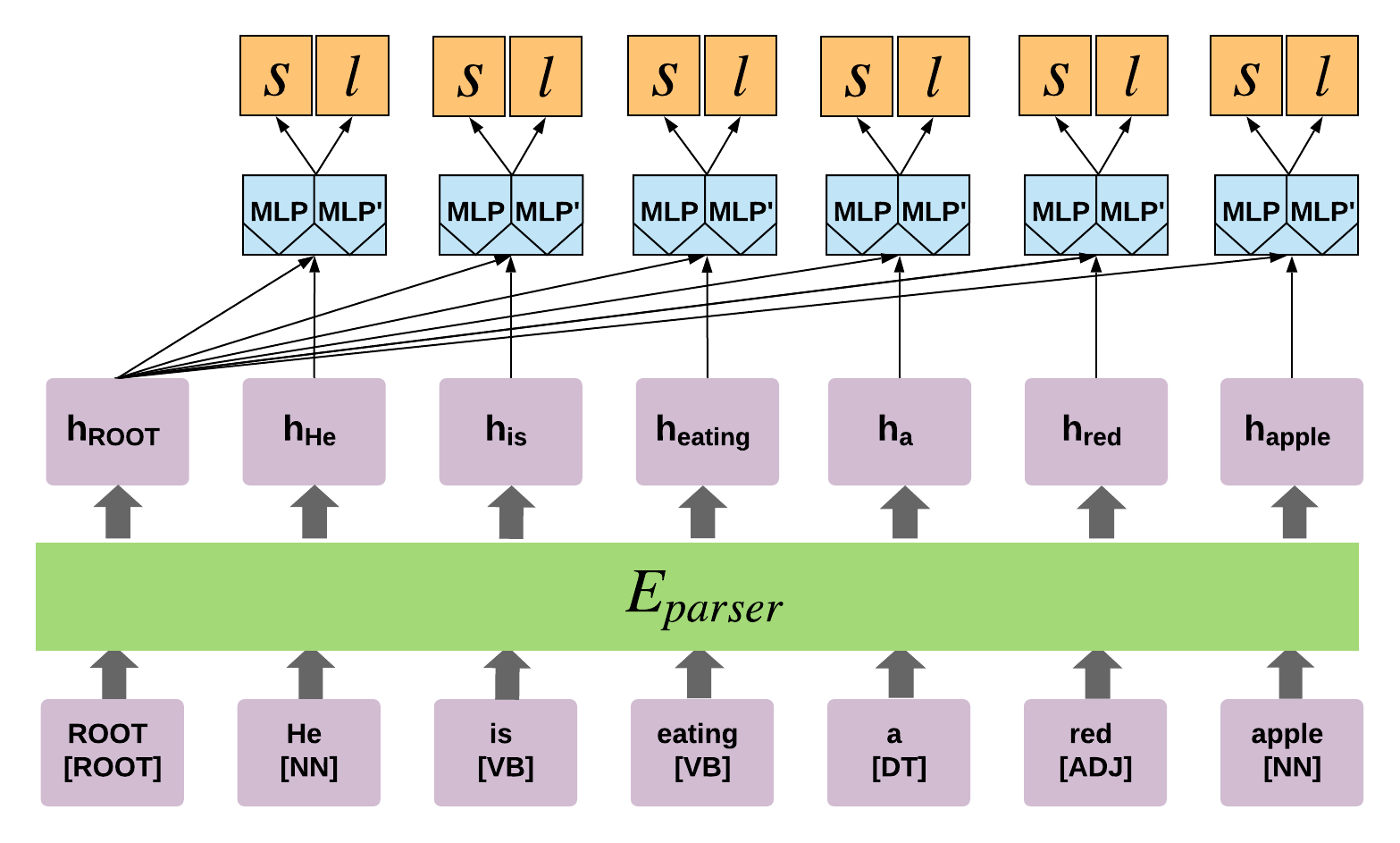}
\caption{The \biaffine\ parser. 
}
\label{fig:sldp_prs}
\end{figure}

A sketch of the parser architecture is provided in Figure \ref{fig:sldp_prs}. The input to the parser is a sentence ($x_1, x_2,\ldots, x_m$) of length $m$. An embedding layer embeds the words into fixed-size vectors ($w_1, w_2,\ldots, w_m$). Additionally, character-level embeddings $c_t^k$ retrieved from a CNN \cite{zhang2015character}, and a POS embedding $p_t$, are concatenated to each word vector. At time $t$, the final input vector $f_t = [w_t;c_t;p_t]$ 
is then fed into a BiLSTM encoder $E_{parser}$ which outputs a hidden representation $h_t$:
\begin{equation}
    h_t = E_{parser}(f_t).
\end{equation}

\par Given the hidden representations of the $i$'th word $h_i$ and the $j$'th word  $h_j$ , the decoder outputs a score $s_{i,j}$, indicating the model belief that the latter should be the head of the former in the dependency tree. More  formally,

\begin{equation}
    s_{i,j} = r_i^TUr_j + w_j^Tr_j,
\end{equation}
where $r_i = MLP(h_i)$, and $U$ and $w_j$ are learned parameters ($MLP$ is a multi-layered perceptron).
\par Similarly, a score $l_{i,j,k}$ is calculated for the $k$'th possible dependency label of the arc $(i,j)$:
\begin{equation}
    l_{i,j,k} = q_i^TU^{'}_{k}q_j + w_{k}^{'T}[q_i;q_j] + \textup{b}^{'}_{k},
\end{equation}

where $q_i = MLP^{'}(h_i)$, and $U^{'}_k$, $w^{'}_k$, and $\textup{b}^{'}_{k}$ are learned parameters.
During training the model aims to maximize the probability of the gold tree:

\begin{equation}
\sum_{i=1}^{m} p(y_i | x_i, \theta) + p(y_i' | x_i,y_i,\theta),
\end{equation}

where $y_i$ is the head of $x_i$,  $y_i'$ is the label of the arc $(x_i,y_i)$, $\theta$ represents the model's parameters, $p(y_i|x_i, \theta) \propto exp(s_{x_i,y_i})$, and $p(y_i'|x_i,y_i,\theta) \propto exp(l_{x_i,y_i,y_i'})$. At test time, the parser runs the MST algorithm \cite{edmonds1967optimum} on the arc scores in order to generate a valid tree.

\section{Deep Contextualized Self-training}
\label{sec:algorithm}

In this section we present our DCST algorithm for dependency parsing (Algorithm \ref{alg:self-training}). As a semi-supervised learning algorithm, DCST assumes a labeled dataset $\mathbf{L}=\{(x^l_i,y^l_i)\}_{i=1}^{{|\mathbf{L}|}}$, consisting of sentences and their gold dependency trees, and an unlabeled dataset $\mathbf{U}=\{x^u_i\}_{i=1}^{{|\mathbf{U}|}}$, consisting of sentences only.

\par
We start (Algorithm \ref{alg:self-training}, step 1) by training the base parser (the \biaffine\ parser in our case) on the labeled dataset $\mathbf{L}$. Once trained, the base parser can output a dependency tree for each of the unlabeled sentences in $\mathbf{U}$ (step 2). We then transform the automatic dependency trees generated for $\mathbf{U}$ into one or more  word-level tagging schemes (step 3). In \S~\ref{sec:cls} we elaborate on this step. Then, we train a BiLSTM sequence tagger to predict the word-level tags of $\mathbf{U}$ (step 4). If the automatic parse trees are transformed to more than one tagging scheme, we train multiple BiLTMs -- one for each scheme. Finally, we construct a new parser by integrating the base parser with the representation BiLSTM(s), and train the final parser on the labeled dataset $\mathbf{L}$ (step 5). In this stage, the base parser parameters are randomly initialized, while the parameters of the representation BiLSTM(s) are initialized to those learned in step 4.

We next discuss the three word-level tagging schemes derived from the dependency trees (step 3), and then the gating mechanism employed in order to compose the hybrid parser (step 5).

\begin{algorithm}[!t]
\caption{\footnotesize Deep Contextualized Self-training (DCST)}\label{alg:self-training}
{\footnotesize
\textbf{Input:} Labeled data $\mathbf{L}$, Unlabeled data $\mathbf{U}$\\
$\textbf{Algorithm:}$
\begin{enumerate}
    \item Train the base parser on $\mathbf{L}$ (\S~\ref{sec:biaffine}).
    \item Parse the sentences of $\mathbf{U}$ with the base parser.
    \item Transform the automatically parsed trees of $\mathbf{U}$ to one or more word-level tagging schemes (\S~\ref{sec:cls}).
    \item Train (a) contextualized embedding model(s) to predict the word-level tagging(s) of $\mathbf{U}$ (\S~\ref{sec:cls}).
    \item Integrate the representation model(s) of step 4 with the base parser, and train the resulting parser on $\mathbf{L}$ (\S~\ref{sec:prs_final}).
\end{enumerate}
}
\end{algorithm}

\subsection{Representation Learning (Steps 3 and 4)}
\label{sec:cls}

In what follows we present the three word-level tagging schemes we consider at step 3 of the DCST algorithm. Transferring the parse trees into tagging schemes is the key for populating information from the original (base) parser on unlabeled data, in a way that can later be re-encoded to the parser through its word embedding layers. The key challenge we face when implementing this idea is the transformation of dependency trees into word level tags that preserve important aspects of the information encoded in the trees. 

We consider tagging schemes that maintain various aspects of the structural information encoded in the tree. Particularly, we start from two tagging schemes that even if fully predicted still leave ambiguity about the actual parse tree: the number of direct dependants each word has and the distance of each word from the root of the tree. We then consider a tagging scheme, referred to as the Relative POS-based scheme, from which the dependency tree can be fully reconstructed. While other tagging schemes can definitely be proposed, we believe that the ones we consider here span a range of possibilities that allows us to explore the validity of our DCST framework.

More specifically, the tagging schemes we consider are defined as follows:

\paragraph{Number of Children}
\label{cls:nos}

Each word is tagged with the number of its children in the dependency tree. We consider only direct children, rather than other descendants, which is equivalent to counting the number of outgoing edges of the word in the tree.

\paragraph{Distance from the Root}
\label{cls:dist}

Each word is tagged with its minimal distance from the root of the tree. For example, if the arc ($ROOT$, $j$) is included in the tree, the distance of the $j$'th word from the ROOT is 1. Likewise, if ($ROOT$, $j$) is not included but ($ROOT$,$i$) and ($i$,$j$) are, then $j$'th distance is 2.

\paragraph{Relative POS-based Encoding}
\label{cls:rpos}

Each word is tagged with its head word according to the relative POS-based scheme \cite{spoustova2010dependency, strzyz2019viable} The head of a word is encoded by a pair $(p, e)\in P\times [-m+1,m-1]$, where $P$ is the set of all possible parts of speech and $m$ is the sentence length. For a positive (negative) number $e$ and a POS $p$, the pair indicates that the head of the represented word is the $e$'th word to its right (left) with the POS tag $p$. To avoid sparsity we coarsen the POS tags related to  nouns, proper names, verbs, adjectives, punctuation-marks and brackets into one tag per category.

Although this word-level tagging scheme was introduced as means of formulating dependency parsing as a sequence tagging task, in practice sequence models trained on this scheme are not competitive with state-of-the-art parsers and often generate invalid tree structures  \cite{strzyz2019viable}. Here we investigate the power of this scheme as part of a self-training algorithm.

\begin{figure}[!t]
\centering
\includegraphics[width=0.99\linewidth]{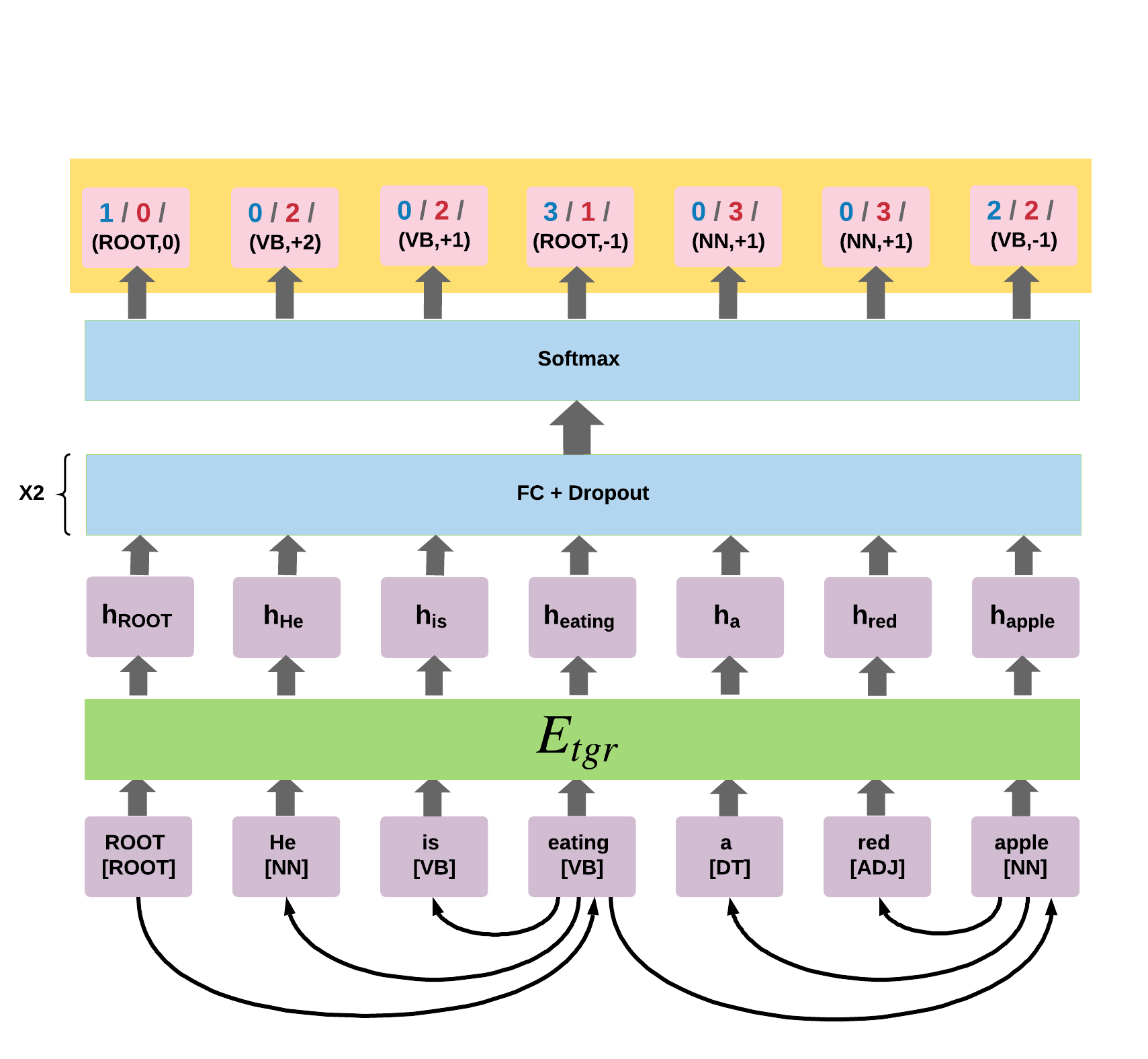}
\caption{The sequence tagger applied to automatically parsed sentences in $\mathbf{U}$ (Algorithm 1, step 4). The tagger predicts for each word its label according to one of the three tagging schemes: Number of Children (blue), Distance from the Root (red), and Relative POS-based Encoding (black). The curved arrows sketch the gold dependency tree from which the word-level tags are derived.}
\label{fig:sldp_cls}
\end{figure}

\paragraph{The Sequence Tagger}

Our goal is to encode the information in the automatically parsed trees into a model that can be integrated with the parser at later stages. This is why we choose to transform the parse trees into word-level tagging schemes that can be learned accurately and efficiently by a sequence tagger. Note that efficiency plays a key role in the lightly-supervised and domain adaptation setups we consider, as large amounts of unlabeled data should compensate for the lack of labeled training data from the target domain. 

We hence choose a simple sequence tagging architecture, depicted in Figure~\ref{fig:sldp_cls}.
The encoder $E_{tgr}$ is a BiLSTM, similarly to $E_{parser}$ of the parser.
The decoder is composed of two fully connected (FC) layers with dropout \cite{srivastava2014dropout} and an exponential linear unit (ELU) activation function \cite{clevert2015fast}, followed by a final softmax layer that outputs the tag probabilities.

\begin{figure}[!t]
\centering
\includegraphics[width=0.99\linewidth]{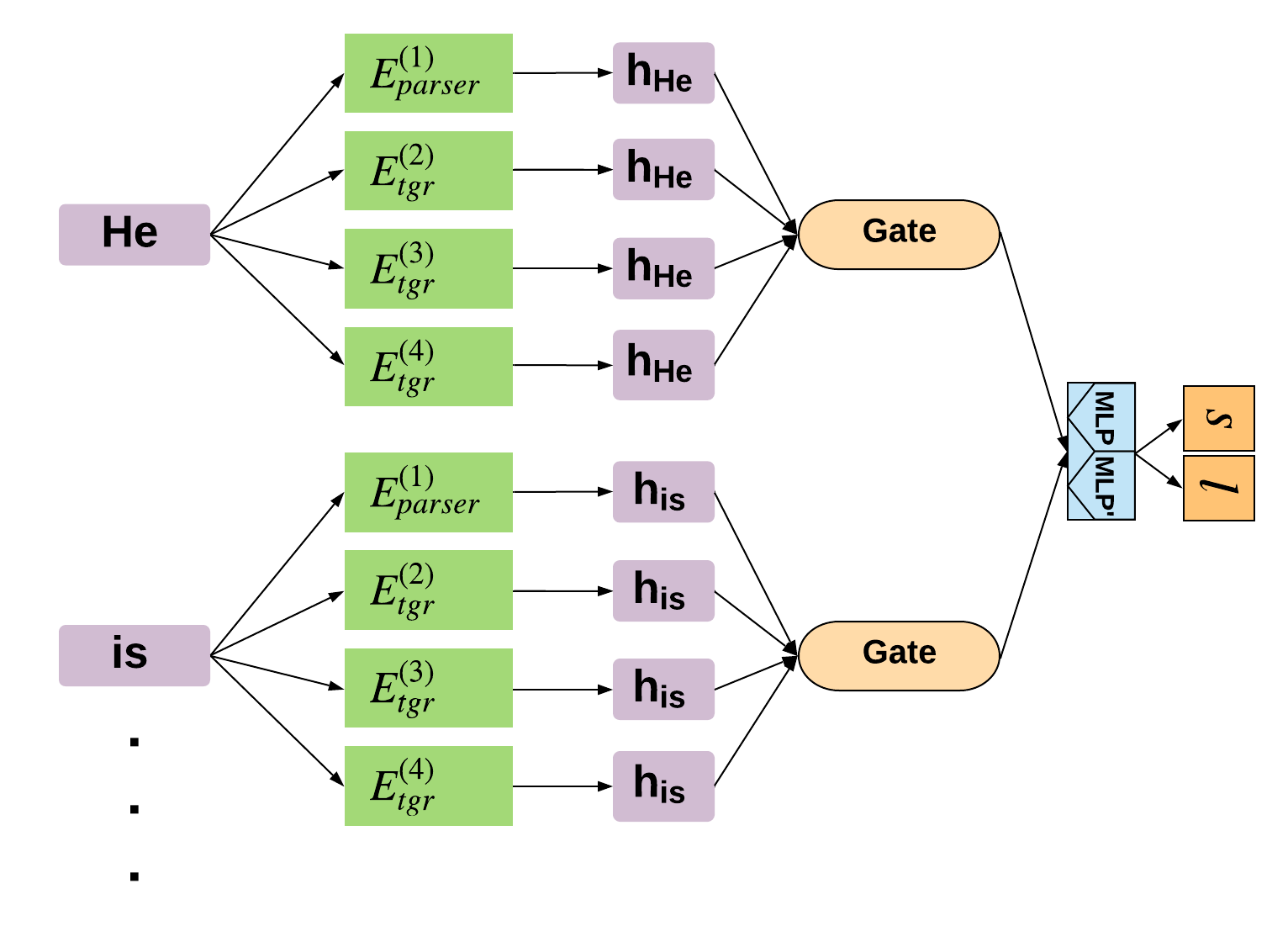}
\caption{An illustration of the hybrid parser with three auxiliary sequence taggers. An input word vector is passed through the parser encoder ($E_{parser}^{(1)}$) and the three pre-trained tagger encoders ($E_{tgr}^{(2)}-E_{tgr}^{(4)}$). The gating mechanism (Gate) computes a weighted average of the hidden vectors. Finally, the output of the gating mechanism is passed to the \biaffine\ decoder to predict the arc and label scores for each word pair.}
\label{fig:sldp_prs_cls_gating}
\end{figure}

\subsection{The Final Hybrid Parser (Step 5)}
\label{sec:prs_final}

In step 5, the final step of Algorithm \ref{alg:self-training}, we integrate the BiLSTM of the sequence tagger, which encodes the information in the automatically generated dependency trees, with the base parser. Importantly, when doing so we initialize the BiLSTM weights to those to which it converged at step 4. The parameters of the base (\biaffine) parser, in contrast, are randomly initialized. The resulting hybrid parser is then trained on the labeled data in $\mathbf{L}$. This way, the final model integrates the information from both $\mathbf{L}$ and the automatic tagging of $\mathbf{U}$, generated in step 2 and 3.

We next describe how the encoders of the sequence tagger and the \biaffine\ parser, $E_{tgr}$ and $E_{parser}$, are integrated through a gating mechanism, similar to that of \citet{sato2017adversarial}.

\paragraph{The Gating Mechanism}

Given an input word vector $f_t$  (\S~\ref{sec:biaffine}), the gating mechanism learns to scale between the BiLSTM encoder of the parser to that of the sequence tagger (Figure \ref{fig:sldp_prs_cls_gating}):

\begin{equation*}
 \begin{gathered}
 a_t = \sigma (W_g[E_{parser}(f_t);E_{tgr}(f_t)] + b_g),  \\
 g_t = a_t \odot E_{parser}(f_t) + (1-a_t) \odot E_{tgr}(f_t). 
 \end{gathered}
\end{equation*}

where $\odot$ is the element-wise product, $\sigma$ is the sigmoid function, and $W_g$ and $b_g$ are the gating mechanism parameters.
The combined vector $g_t$ is then fed to the parser's decoder.

\paragraph{Extension to $\boldsymbol{n} \geq 2$ Sequence Taggers}

We can naturally extend our hybrid parser to support $n$ auxiliary taggers (see again Figure \ref{fig:sldp_prs_cls_gating}). Given $n$ taggers trained on $n$ different tagging schemes, 
the gating mechanism is modified to be:
\begin{equation*}
  \begin{gathered}
 s_t^{(i)} =\\ W_g^{(i)}[E_{parser}^{(1)}(f_t);E_{tgr}^{(2)}(f_t);\ldots ;E_{tgr}^{(n+1)}(f_t)] + b_g^{(i)} , \\
 a_t^{(i)} = \frac{exp(s_t^{(i)})}{\sum_{j=1}^{n+1}exp(s_t^{(j)})}, \\
 g_t =  a_t^{(1)} \odot E_{parser}^{(1)}(f_t) + \sum_{i=2}^{n+1} a_t^{(i)} \odot E_{tgr}^{(i)}(f_t). 
  \end{gathered}
\end{equation*}
This extension provides a richer representation of the automatic tree structures, as every tagging scheme captures a different aspect of the trees. Indeed, in most of our experiments, when integrating the base parser with our three proposed schemes, the resulting model was superior to models that consider a single tagging scheme.

\section{Evaluation Setups}
\label{sec:setups}

This paper focuses on exploiting unlabeled data in order to improve the accuracy of a supervised parser. We expect this approach to be most useful when the parser does not have sufficient labeled data for training, or when the labeled training data do not come from the same distribution as the test data. We hence focus on two setups: 

\paragraph{The Lightly Supervised In-domain Setup}

In this setup we are given a small labeled dataset $\mathbf{L}=\{(x^l_i,y^l_i)\}_{i=1}^{{|\mathbf{L}|}}$ of sentences and their gold dependency trees and a large unlabeled dataset $\mathbf{U}=\{(x^u_i)\}_{i=1}^{{|\mathbf{U}|}}$ of sentences coming from the same domain, where $|\mathbf{L}| \ll |\mathbf{U}|$. Our goal is to parse sentences from the domain of $\mathbf{L}$ and $\mathbf{U}$.

\paragraph{The Unsupervised Domain Adaptation Setup}

In this setup we are given a labeled source domain dataset $\mathbf{L}=\{(x^l_i,y^l_i)\}_{i=1}^{{|\mathbf{L}|}}$ of sentences and their gold dependency trees, and an unlabeled dataset $\mathbf{U}=\{(x^u_i)\}_{i=1}^{{|\mathbf{U}|}}$ of sentences from a different target domain. Unlike the lightly-supervised setup, here $\mathbf{L}$ may be large enough to train a high quality parser as long as the training and test sets come from the same domain. However, our goal here is to parse sentences from the target domain.







\com{
\begin{table*}[!ht]
\def\arraystretch{0.9}
\scalebox{0.67}{
\centering
\begin{tabular}{cllllllllllllll}
&
\multicolumn{2}{c}{bc} &
\multicolumn{2}{c}{bn} &
\multicolumn{2}{c}{mz} &
\multicolumn{2}{c}{nw} &
\multicolumn{2}{c}{pt} &
\multicolumn{2}{c}{tc} &
\multicolumn{2}{c}{wb} \\
\cmidrule(lr){2-3}
\cmidrule(lr){4-5}
\cmidrule(lr){6-7}
\cmidrule(lr){8-9}
\cmidrule(lr){10-11}
\cmidrule(lr){12-13}
\cmidrule(lr){14-15}
{\bf Model} & UAS & LAS & UAS & LAS & UAS & LAS & UAS & LAS & UAS & LAS & UAS & LAS & UAS & LAS \\
\cmidrule(lr){2-3}
\cmidrule(lr){4-5}
\cmidrule(lr){6-7}
\cmidrule(lr){8-9}
\cmidrule(lr){10-11}
\cmidrule(lr){12-13}
\cmidrule(lr){14-15}
Base
& 74.54	& 70.77
& 80.57	& 77.63
& 81.47	& 78.41
& 80.40	& 77.56
& 86.95	& 83.86
& 72.15	& 68.34
& 78.74	& 73.24 \\ 
Base+RG
& 77.10	& 73.45
& 81.90	& 79.06
& 83.02	& 80.29
& 81.80	& 79.24
& 88.13	& 85.42
& 73.87	& 69.97
& 78.93	& 75.37  \\
DCST-LM 
& 75.94	& 72.33
& 80.01	& 76.96
& 82.50	& 79.53
& 80.33	& 77.46
& 87.53	& 84.56
& 72.16	& 68.30
& 77.09	& 73.49\\
Self-Training
& 74.64	& 71.18
& 82.35	& 79.75
& 83.44	& 80.86
& 81.93	& 79.43
& 87.50	& 84.52
& 69.70	& 66.62
& 79.18	& 75.86  \\ 
CVT
& 78.47	& 73.54
& 82.76	& 78.19
& 82.90 & 78.56
& \underline{\textbf{85.55}}	& \underline{\textbf{82.30}}
& \underline{\textbf{90.36}}	& \underline{\textbf{87.05}}
& 75.36	& 69.96
& 78.03	& 73.10  \\ 
\midrule
DCST-NC   
& 78.21	& 74.62
& 82.32	& 79.52
& 83.52	& 80.61
& 81.95	& 79.17
& 88.83	& 85.62
& 75.35	& 71.05
& 78.76	& 75.10 \\
DCST-DR 
& 78.61	& 74.80
& 83.32	& 80.26
& 84.27	& 81.15
& 82.67	& 79.74
& 88.90	& 85.66
& 75.05	& 70.82
& 79.80	& 76.12 \\ 
DCST-RPE
& 78.70	& 75.11
& 83.07	& 80.41
& 84.16	& 81.62
& 83.02	& 80.45
& 88.95	& 85.96
& 75.35	& 71.06
& 80.25	& 76.91
\\ 
DCST-ENS
& \underline{\textbf{78.95}}	& \underline{\textbf{75.43}}
& \underline{\textbf{83.52}}	& \underline{\textbf{80.93}}
& \underline{\textbf{84.67}}	& \underline{\textbf{81.99}}
& 82.89	& 80.41
& 89.38	& 86.47
& \textbf{76.47}	& \underline{\textbf{72.54}}
& \underline{\textbf{80.52}}	& \underline{\textbf{77.32}}
\\
\end{tabular}}
\caption{Lightly supervised OntoNotes results with 500 training sentences. \com{Underscored results are significant compared to the highest scoring baseline, according to t-test with $p < 0.05$.}}
\label{table:onto_500}
\end{table*}
}

\section{Experiments}
\label{sec:experiments}

We experiment with the task of dependency parsing, in two setups: (a) lightly supervised in-domain and (b) unsupervised domain adaptation. 

\paragraph{Data}

We consider two datasets: \textbf{(a)} The English OntoNotes 5.0 \cite{hovy2006ontonotes} corpus. This corpus consists of text from 7 domains: broadcast conversation (bc: 11877 training, 2115 development and 2209 test sentences), broadcast news (bn: 10681, 1293, 1355), magazine (mz: 6771, 640, 778), news (nw: 34967, 5894, 2325), bible (pt: 21518, 1778, 1867), telephone conversation (tc: 12889, 1632, 1364) and web (wb: 15639, 2264, 1683).\footnote{We removed wb test set sentences where all words are POS tagged with "XX".} The corpus is annotated with constituency parse trees and POS tags, as well as other labels that we do not use in our experiments. The constituency trees were converted to dependency trees using the Elitcloud conversion tool.\footnote{\url{https://github.com/elitcloud/elit-java}.} In the lightly supervised setup we experiment with each domain separately. We further utilize this corpus in our domain adaptation experiments. \textbf{(b)} The Universal Dependencies (UD) dataset \cite{mcdonald2013universal, nivre2016universal, nivre2018universal}. This corpus contains more than 100 corpora of over 70 languages, annotated with dependency trees and universal POS tags. For the lightly supervised setup we chose 10 low-resource languages that have no more than 10K training sentences: Old Church Slavonic (cu), Danish (da), Persian (fa), Indonesian (id), Latvian (lv), Slovenian (sl), Swedish (sv), Turkish (tr), Urdu (ur) and Vietnamese (vi), and performed mono-lingual experiments with each.\footnote{In case a language has multiple corpora, our training, development and test sets are concatenations of the corresponding sets in these corpora.} For the domain adaptation setup we experiment with 5 languages, considering two corpora from different domains for each: Czech (cs\_fictree: fiction, cs\_pdt: news and science), Galician (gl\_ctg: science and legal, gl\_treegal: news), Italian (it\_isdt: legal, news and wiki, it\_postwita: social media), Romanian (ro\_nonstandard: poetry and bible, ro\_rrt: news, literature, science, legal and wiki) and Swedish (sv\_lines: literature and politics, sv\_talbanken: news and textbooks).

\paragraph{Training Setups}

For the lightly supervised setup we performed experiments with the 7 OntoNotes domains and the 10 UD corpora, for a total of 17 in-domain setups. For each setup we consider three settings that differ from each other in the size of the randomly selected labeled training and development sets: 100, 500 or 1000.\footnote{In languages where the development set was smaller than 1000 sentences we used the entire development set.} We use the original test sets for evaluation, and the remaining training and development sentences as unlabeled data.

For the English unsupervised domain adaptation setup, we consider the news (nw) section of OntoNotes 5.0 as the source domain, and the remaining sections as the target domains. The nw training and development sets are used for the training and development of the parser, while the unlabeled versions of the target domain training and development sets are used for training and development of the representation model. The final model is evaluated on the target domain test set.

Similarly, for unsupervised domain adaptation with the UD languages, we consider within each language one corpus as the source domain and the other as the target domain, and apply the same train/development/test splits as above. For each language we run two experiments, differing in which of the two corpora is considered the source and which is considered the target.

For all domain adaptation experiments, when training the final hybrid parser (Figure \ref{fig:sldp_prs_cls_gating}) we sometimes found it useful to keep the parameters of the BiLSTM tagger(s) fixed in order to avoid an overfitting of the final parser to the source domain. We treat the decision of whether or not to keep the parameters of the tagger(s) fixed as a hyper-parameter of the DCST models and tune it on the development data.


We measure parsing accuracy with the standard Unlabeled and Labeled Attachment Scores (UAS and LAS), and
measure statistical significance with the t-test (following \citet{dror-etal-2018-hitchhikers}).

\paragraph{Models and Baselines}

We consider four variants of our DCST algorithm, differing on the word tagging scheme on which the BiLSTM of step 4 is trained (\S~\ref{sec:cls}): \textbf{DCST-NC}: with the Number of Children scheme, \textbf{DCST-DR}: with the Distance from the Root scheme, \textbf{DCST-RPE}: with the Relative POS-based Encoding scheme and \textbf{DCST-ENS} where the parser is integrated with three BiLSTMs, one for each scheme (where ENS stands for ensemble) (\S~\ref{sec:prs_final}).

To put the results of our DCST algorithm in context, we compare its performance to the following baselines. \textbf{Base}: the \biaffine\ parser (\S~\ref{sec:biaffine}), trained on the labeled training data. \textbf{Base-FS}: the \biaffine\ parser (\S~\ref{sec:biaffine}), trained on all the labeled data available in the full training set of the corpus. In the domain adaptation setups Base-FS is trained on the entire training set of the target domain. This baseline can be thought of as an upper bound on the results of a lightly-supervised learning or domain-adaptation method. \textbf{Base + Random Gating (RG)}: a randomly initialized BiLSTM is integrated to the \biaffine\ parser through the gating mechanism, and the resulting model is trained on the labeled training data. We compare to this baseline in order to quantify the effect of the added parameters of the BiLSTM and the gating mechanism, when this mechanism does not inject any information from unlabeled data. 
\textbf{Self-training}: the traditional self-training procedure. We first train the Base parser on the labeled training data, then use the trained parser to parse the unlabeled data, and finally re-train the Base parser on both the manual and automatic trees.

We would also like to test the value of training a representation model to predict the dependency labeling schemes of \S~\ref{sec:cls}, in comparison to the now standard pre-training with a language modeling objective. Hence, we experiment with a variant of DCST where the BiLSTM of step 4  is trained as a language model (\textbf{DCST-LM}). 
Finally, we compare to the cross-view training algorithm (\textbf{CVT}) \cite{clark2018semi}, that was developed for semi-supervised learning with DNNs.
\footnote{\url{https://github.com/tensorflow/models/tree/master/research/cvt_text}.}

\paragraph{Hyper-parameters}

We employ the \biaffine\ parser implementation of  \citet{ma2018stack}. \footnote{\url{https://github.com/XuezheMax/NeuroNLP2}.} We consider the following hyper-parameters for the parser and the sequence tagger: 100 epochs with an early stopping criterion according to the development set, the ADAM optimizer \cite{kingma2014adam}, a batch size of 16, a learning rate of 0.002 and dropout probabilities of 0.33. 

The 3-layer stacked  BiLSTMs of the parser and the sequence tagger generate hidden representations of size 1024. The fully connected layers of the tagger are of size 128 (first layer) and 64 (second layer). All other parser hyper-parameters are identical to those of the original implementation.

We employ 300-dimensional pre-trained word embeddings: GloVe \cite{pennington2014glove} \footnote{\url{http://nlp.stanford.edu/data/glove.840B.300d.zip}.} for English and FastText \cite{grave2018learning} \footnote{\url{https://fasttext.cc/docs/en/crawl-vectors.html}.} for the UD languages. Character and POS embeddings are 100-dimensional and are initialized to random normal vectors. 
%
\textbf{CVT} is run for 15K gradient update steps.


\section{Results}
\label{sec:results}



\begin{table*}[!ht]
\def\arraystretch{0.9}
\scalebox{0.75}{
\centering
\begin{tabular}{cllllllllllllll}
&
\multicolumn{2}{c}{bc} &
\multicolumn{2}{c}{bn} &
\multicolumn{2}{c}{mz} &
\multicolumn{2}{c}{nw} &
\multicolumn{2}{c}{pt} &
\multicolumn{2}{c}{tc} &
\multicolumn{2}{c}{wb} \\
\cmidrule(lr){2-3}
\cmidrule(lr){4-5}
\cmidrule(lr){6-7}
\cmidrule(lr){8-9}
\cmidrule(lr){10-11}
\cmidrule(lr){12-13}
\cmidrule(lr){14-15}
{\bf Model} & UAS & LAS & UAS & LAS & UAS & LAS & UAS & LAS & UAS & LAS & UAS & LAS & UAS & LAS \\
\cmidrule(lr){2-3}
\cmidrule(lr){4-5}
\cmidrule(lr){6-7}
\cmidrule(lr){8-9}
\cmidrule(lr){10-11}
\cmidrule(lr){12-13}
\cmidrule(lr){14-15}
Base
& 74.54	& 70.77
& 80.57	& 77.63
& 81.47	& 78.41
& 80.40	& 77.56
& 86.95	& 83.86
& 72.15	& 68.34
& 78.74	& 73.24 \\ 
Base+RG
& 77.10	& 73.45
& 81.90	& 79.06
& 83.02	& 80.29
& 81.80	& 79.24
& 88.13	& 85.42
& 73.87	& 69.97
& 78.93	& 75.37  \\
DCST-LM 
& 75.94	& 72.33
& 80.01	& 76.96
& 82.50	& 79.53
& 80.33	& 77.57
& 87.53	& 84.56
& 72.16	& 68.30
& 77.09	& 73.49\\
Self-Training
& 74.64	& 71.18
& 82.35	& 79.75
& 83.44	& 80.86
& 81.93	& 79.43
& 87.50	& 84.52
& 69.70	& 66.62
& 79.18	& 75.86  \\ 
CVT
& 78.47	& 73.54
& 82.76	& 78.19
& 82.90 & 78.56
& \underline{\textbf{85.55}}	& \underline{\textbf{82.30}}
& \underline{\textbf{90.36}}	& \underline{\textbf{87.05}}
& 75.36	& 69.96
& 78.03	& 73.10  \\ 
\midrule
DCST-NC   
& 78.21	& 74.62
& 82.32	& 79.52
& 83.52	& 80.61
& 81.95	& 79.17
& 88.83	& 85.62
& 75.35	& 71.05
& 78.76	& 75.10 \\
DCST-DR 
& 78.61	& 74.80
& 83.32	& 80.26
& 84.27	& 81.15
& 82.67	& 79.74
& 88.90	& 85.66
& 75.05	& 70.82
& 79.80	& 76.12 \\ 
DCST-RPE
& 78.70	& 75.11
& 83.07	& 80.41
& 84.16	& 81.62
& 83.02	& 80.45
& 88.95	& 85.96
& 75.35	& 71.06
& 80.25	& 76.91
\\ 
DCST-ENS
& \underline{\textbf{78.95}}	& \underline{\textbf{75.43}}
& \underline{\textbf{83.52}}	& \underline{\textbf{80.93}}
& \underline{\textbf{84.67}}	& \underline{\textbf{81.99}}
& 82.89	& 80.41
& 89.38	& 86.47
& \textbf{76.47}	& \underline{\textbf{72.54}}
& \underline{\textbf{80.52}}	& \underline{\textbf{77.32}} \\
\midrule
Base-FS 
& 86.23	& 84.49
& 89.41	& 88.17
& 89.19	& 87.80
& 89.29	& 88.01
& 94.08	& 92.83
& 77.12	& 75.36
& 87.23	& 85.56
\end{tabular}}
\caption{Lightly supervised OntoNotes results with 500 training sentences. \com{The best result for each setup and measure is highlighted in bold.} Base-FS is an upper bound. \com{that puts our results in context.}  \com{Underscored results are significant compared to the highest scoring baseline, according to t-test with $p < 0.05$.}}
\label{table:onto_500}
\end{table*}

\begin{table*}[!ht]
\def\arraystretch{0.9}
\scalebox{0.56}{
\centering
\begin{tabular}{c llllllllllllllllllll}
{}    &
\multicolumn{2}{c}{cu} &
\multicolumn{2}{c}{da} &
\multicolumn{2}{c}{fa} &
\multicolumn{2}{c}{id} &
\multicolumn{2}{c}{lv} &
\multicolumn{2}{c}{sl} &
\multicolumn{2}{c}{sv} &
\multicolumn{2}{c}{tr} &
\multicolumn{2}{c}{ur} &
\multicolumn{2}{c}{vi} \\
\cmidrule(lr){2-3}
\cmidrule(lr){4-5}
\cmidrule(lr){6-7}
\cmidrule(lr){8-9}
\cmidrule(lr){10-11}
\cmidrule(lr){12-13}
\cmidrule(lr){14-15}
\cmidrule(lr){16-17}
\cmidrule(lr){18-19}
\cmidrule(lr){20-21}

{\bf Model} & UAS & LAS & UAS & LAS & UAS & LAS & UAS & LAS & UAS & LAS & UAS & LAS & UAS & LAS & UAS & LAS & UAS & LAS & UAS & LAS   \\
\cmidrule(lr){2-3}
\cmidrule(lr){4-5}
\cmidrule(lr){6-7}
\cmidrule(lr){8-9}
\cmidrule(lr){10-11}
\cmidrule(lr){12-13}
\cmidrule(lr){14-15}
\cmidrule(lr){16-17}
\cmidrule(lr){18-19}
\cmidrule(lr){20-21}
Base
& 75.87	& 67.25
& 78.13	& 74.16
& 82.54	& 78.59
& 72.57	& 57.25
& 72.81	& 65.66
& 76.00	& 69.28
& 78.58	& 72.78
& 56.07	& 39.37
& 84.49	& 78.10
& 67.18	& 62.51 \\
Base+RG
& 77.98	& 69.01
& 80.21	& 76.11
& 84.74	& 80.83
& 73.18	& 57.56
& 74.51	& 67.60
& 78.18	& 71.27
& 79.90	& 73.70
& 58.42	& 40.32
& 86.18	& 79.65
& 68.75	& 64.64  \\
DCST-LM 
& 77.67	& 68.90
& 80.23	& 76.06
& 83.92	& 79.89
& 72.61	& 57.36
& 73.89	& 66.59
& 76.90	& 70.12
& 78.73	& 72.51
& 57.33	& 39.27
& 85.78	& 79.27
& 69.11	& 65.09 \\
Self-Training
& 75.19	& 68.07
& 79.76	& 75.92
& 85.04	& 81.05
& 74.07	& 58.73
& 74.79	& 68.22
& 77.71	& 71.33
& 79.72	& 74.12
& 57.34	& 40.06
& 85.63	& 79.51
& 68.24	& 63.96  \\ 
CVT
& 61.57	& 45.60
& 72.77	& 66.93
& 81.08	& 74.32
& 72.51	& 54.94
& 68.90	& 57.36
& 67.89	& 59.79
& 77.08	& 69.60
& 53.17	& 32.95
& 81.49	& 72.72
& 60.84	& 50.98 \\
\midrule
DCST-NC   
& 78.85	& 69.75
& 81.23	& 76.70
& 85.94	& 81.85
& 74.18	& 58.63
& 76.19	& 68.73
& 79.26	& 72.72
& 81.05	& 75.09
& 58.17	& 39.95
& 86.17	& 79.91
& 69.93	& 65.91 \\
DCST-DR 
& 79.31	& 70.20
& 81.30	& 76.81
& 86.20	& 82.14
& \underline{\textbf{74.56}}	& 58.92
& 76.99	& 69.24
& 80.34	& 73.35
& 81.40	& 75.41
& 58.30	& 40.25
& 86.19	& 79.68
& 69.46	& 65.65 \\ 
DCST-RPE
& \underline{\textbf{80.57}}	& \underline{\textbf{71.83}}
& 81.48	& 77.45
& 86.82	& 82.69
& \underline{\textbf{74.56}}	& \underline{\textbf{59.19}}
& 77.45	& \underline{\textbf{70.38}}
& 80.45	& 74.13
& 81.95	& 75.98
& 59.49	& 41.45
& 86.86	& \underline{\textbf{80.92}}
& 70.23	& 66.26 \\ 
DCST-ENS
& 80.55	& 71.79
& \underline{\textbf{82.07}}	& \underline{\textbf{78.04}}
& \underline{\textbf{87.02}}	& \underline{\textbf{83.13}}
& 74.47	& 59.13
& \underline{\textbf{77.63}}	& 70.36
& \underline{\textbf{80.68}}	& \underline{\textbf{74.32}}
& \underline{\textbf{82.40}}	& \underline{\textbf{76.61}}
& \underline{\textbf{59.60}}	& \underline{\textbf{41.72}}
& \underline{\textbf{86.96}}	& 80.85
& \textbf{70.37}	& \underline{\textbf{66.88}} \\
\midrule
Base-FS
& 86.13	& 81.46
& 85.55	& 82.93
& 91.06	& 88.12
& 77.42	& 62.31
& 85.02	& 81.59
& 86.04	& 82.22
& 85.18	& 81.36
& 62.21	& 46.23
& 89.84	& 85.12
& 73.26	& 69.69 \\
\end{tabular}}

\caption{Lightly supervised UD results with 500 training sentences. \com{The best result for each setup and measure is highlighted in bold.} Base-FS is an upper bound. \com{that puts our results in context.} \com{Underscored results are significant compared to the highest scoring baseline, according to t-test with $p < 0.05$.} 
}
\label{table:ud_500}
\end{table*}

Table \ref{table:onto_500} presents the lightly supervised OntoNotes results when training with  500 labeled sentences, while Table \ref{table:ud_500} presents the UD results in the same setup. Tables \ref{table:onto_da} and \ref{table:ud_da} report domain adaptation results for the 6 OntoNotes and 10 UD target domains, respectively. Underscored results are significant compared to the highest scoring baseline, based on t-test with $p < 0.05$.\footnote{For this comparison, Base-FS is not considered a baseline, but an upper bound.}

\paragraph{DCST with Syntactic Self-training}

DCST-ENS, our model that integrates all three syntactically self-trained BiLSTMs, is clearly the best model. In the lightly supervised setup, it performs best on 5 of 7 OntoNotes domains and on 8 of 10 UD corpora (with the UAS measure). In the cases where DCST-ENS is not the best performing model, it is the second or third best model. In the English and multilingual domain adaptation setups, DCST-ENS is clearly the best performing model, where in only 2 multilingual target domains it is second.

Moreover, DCST-NC, DCST-DR and DCST-RPE, that consider only one syntactic scheme, also excel in the lightly supervised setup. They outperform all the baselines (models presented above the top separating lines in the tables) in the UD experiments, and DCST-RPE and DCST-DR outperform all the baselines in 5 of 7 Ontonotes domains (with the LAS measure). In the domain adaptation setup, however, they are on par with the strongest baselines, which indicates the importance of exploiting the information in all three schemes in this setup (results are not shown in Tables \ref{table:onto_da} and \ref{table:ud_da} in order to save space). 

Note, that with few exceptions, DCST-NC is the least effective method among the syntactically self-trained DCST alternatives. This indicates that encoding the number of children each word has in the dependency tree is not a sufficiently informative view of the tree.

\paragraph{Comparison to Baselines}

The CVT algorithm performs quite well in the English OntoNotes lightly supervised setup -- it is the best performing model on two domains (nw and pt) and the  best baseline for three other domains when considering the UAS measure (bc, bn and tc). However, its performance substantially degrades in  domain adaptation. 
Particularly, in 5 out of 6 OntoNotes setups and in 9 out of 10 UD setups it is the worst performing model. Moreover, CVT is the worst performing model in the lightly supervised multilingual setup. 

Overall, this recently proposed model that demonstrated strong results across several NLP tasks, does not rival our DCST models with syntactic self-training in our experimental tasks. Notice that \newcite{clark2018semi} did not experiment in domain adaptation setups and did not consider languages other than English. Our results suggest that in these cases DCST with syntactic self-training is a better alternative. 

\begin{table}[!ht]
\def\arraystretch{0.9}
\scalebox{0.75}{
\centering
\begin{tabular}{c cccccc}
&
bc &
bn &
mz &
pt &
tc &
wb \\
\cmidrule(lr){2-3}
\cmidrule(lr){4-5}
\cmidrule(lr){6-7}
{\bf Model} & LAS & LAS & LAS & LAS & LAS & LAS \\
\cmidrule(lr){2-3}
\cmidrule(lr){4-5}
\cmidrule(lr){6-7}
Base
& 81.60
& 85.17
& 85.48
& 87.70
& 75.46
& 83.85 \\
Base+RG
& 82.51
& 85.36
& 85.77
& 88.34
& 75.68
& 84.34 \\
DCST-LM 
& 82.48
& 85.77
& 86.28
& 89.28
& 75.72
& 84.34 \\
Self-Training
& 80.61
& 84.52
& 85.38
& 87.69
& 73.62
& 82.82 \\
CVT
& 74.81
& 84.90
& 84.49
& 85.71
& 72.10
& 82.31 \\
\midrule
DCST-ENS
& \underline{\textbf{85.96}}
& \underline{\textbf{88.02}}
& \underline{\textbf{88.55}}
& \underline{\textbf{91.62}}
& \underline{\textbf{79.97}}
& \underline{\textbf{87.38}} \\
\midrule
Base-FS 
& 84.49
& 88.17
& 87.80
& 92.83
& 75.36
& 85.56 \\
\end{tabular}}
\caption{Unsupervised Domain adaptation OntoNotes results. \com{The best result for each setup is highlighted in bold.} Base-FS is an upper bound. \com{that puts our results in context.}   \com{on OntoNotes 5.0. Underscored results are significant compared to the baseline with the highest score, according to t-test with $p < 0.05$.}}
\label{table:onto_da}
\end{table}

\begin{table*}[!ht]
\def\arraystretch{0.9}
\scalebox{0.75}{
\centering
\begin{tabular}{c cccccccccc}
&
cs\_fictree &
cs\_pdt &
gl\_ctg &
gl\_treegal &
it\_isdt &
it\_postwita &
ro\_nonstandard &
ro\_rrt &
sv\_lines &
sv\_talbanken \\
\cmidrule(lr){2-3}
\cmidrule(lr){4-5}
\cmidrule(lr){6-7}
\cmidrule(lr){8-9}
\cmidrule(lr){10-11}
{\bf Model} & LAS & LAS & LAS & LAS & LAS & LAS & LAS & LAS & LAS & LAS \\
\cmidrule(lr){2-3}
\cmidrule(lr){4-5}
\cmidrule(lr){6-7}
\cmidrule(lr){8-9}
\cmidrule(lr){10-11}
Base
& 69.92
& 81.83
& 59.05
& 60.31
& 67.82
& 80.72
& 65.03
& 62.75
& 77.08
& 77.93 \\
Base+RG
& 73.12
& 80.86
& 58.97
& 60.52
& 67.54
& 80.36
& 65.93
& 61.50
& 77.58
& 78.04 \\
DCST-LM 
& 73.59
& 83.33
& 59.41
& 60.54
& 67.52
& 80.95
& 65.19
& 62.46
& 77.40
& 77.62 \\
Self-Training
& 69.50
& 81.53
& 59.67
&  \textbf{61.41}
& 68.02
& 82.01
& 66.47
& \textbf{63.84}
& 77.60
& 77.64 \\
CVT
& 59.77
& 81.53
& 51.12
& 50.31
& 58.60
& 70.07
& 50.82
& 45.15
& 45.25
& 62.87 \\
\midrule
DCST-ENS
& \underline{\textbf{75.28}}
& \underline{\textbf{86.50}}
& \textbf{59.75}
& 60.98
& \underline{\textbf{69.13}}
& \underline{\textbf{83.06}}
& \underline{\textbf{67.65}}
& 63.46
& \textbf{77.86}
& \underline{\textbf{78.97}} \\
\midrule
Base-FS 
& 84.46
& 83.70
& 84.44
& 78.09
& 90.02
& 81.22
& 81.71
& 84.99
& 82.43
& 86.67 \\
\end{tabular}}
\caption{Unsupervised Domain adaptation UD results. \com{The best result for each setup is highlighted in bold.} Base-FS is an upper bound.\com{ that puts our results in context.} }
\label{table:ud_da}
\end{table*}

We next evaluate the impact of the different components of our model. First, comparison with DCST-LM -- the version of our model where the syntactically self-trained BiLSTM is replaced with a BiLSTM trained on the same unlabeled data but with a language modeling objective, allows us to evaluate the importance of the self-generated syntactic signal. The results are conclusive: in all our four setups - English and multilingual lightly-supervised, and English and multilingual domain adaptation, DCST-LM is outperformed by DCST-ENS that considers all three self-trained BiLSTMs. DCST-LM is also consistently outperformed by DCST-RPE, DCST-DR and DCST-NC that consider only one syntactic annotation scheme, except from a few English lightly-supervised cases where it outperforms DCST-NC by a very small margin. Syntactic self-supervision hence provides better means of exploiting the unlabeled data, compared to the standard language modeling alternative.

Another question is whether the BiLSTM models should be trained at all. Indeed, in recent papers untrained LSTMs with random weights substantially enhanced model performance \cite{wang2019can, zhang-bowman-2018-language,Tenney:2019,Wieting:19}.

Our results lead to two conclusions. Firstly, Base+RG, the model that is identical to the syntactically trained DCST except that the \biaffine\ parser is integrated with a randomly initialized BiLSTM through our gating mechanism, is consistently outperformed by all our syntactically self-trained DCST models, with very few exceptions. 
Secondly, in line with the conclusions of the aforementioned papers, Base+RG is one of the strongest baselines in our experiments. Perhaps most importantly, in most experiments this model outperforms the Base parser -- indicating the positive impact of the randomly initialized representation models. Moreover, it is the strongest baseline in 2 English domain adaptation setups and in 5 of 10 languages in the lightly-supervised multilingual experiments (considering the UAS measure), and is the second-best baseline in 5 out of 7 English lightly-supervised setups (again considering the UAS measure). The growing evidence for the positive impact of such randomly initialized models should motivate further investigation of the mechanism that underlies their success.

Finally, our results demonstrate the limited power of traditional self-training: In English domain adaptation it harms or does not improve the Base parser; in multilingual domain adaptation it is the best model in 2 cases; and it is the best baseline in 2 of the 7 English lightly-supervised setups and in 3 of the 10 multilingual lightly-supervised setups. This supports our motivation to propose an improved self-training framework.

\section{Ablation Analysis and Discussion}
\label{sec:ablation}


\paragraph{Impact of Training set Size}

\begin{figure}
\centering
\includegraphics[width=1.0\linewidth]{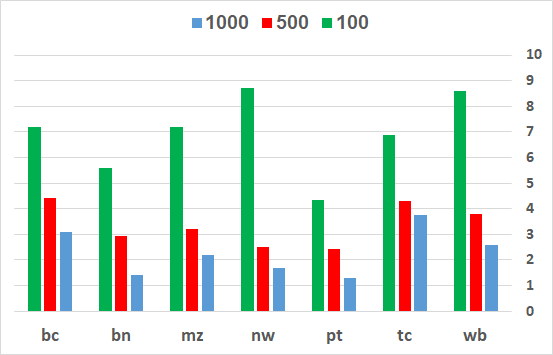}
\caption{UAS gap between DCST-ENS and the Base parser, as a function of the training set size (100/500/1000), across OntoNotes domains.}
\label{fig:gap_bar_chart}
\end{figure}

Figure~\ref{fig:gap_bar_chart} presents the impact of the DCST-ENS method on the \biaffine\ parser, in the seven lightly-supervised English setups, as a function of the labeled training set size of the parser. Clearly, the positive impact is substantially stronger for smaller training sets. Particularly, when the parser is trained with 100 sentences (the green bar) the improvement is higher than 5 UAS points in 6 of 7 cases, among which in 2 (nw and wb) it is higher than 8 UAS points. For 500 training sentences the performance gap drops to 2-4 UAS points, while for 1000 training sentences it is 1-3 points.

This pattern is in line with previous literature on the impact of training methods designed for the lightly-supervised setup, and particularly for self-training when applied to constituency parsing \cite{reichart2007self}. We note that many research papers failed to improve dependency parsing with traditional self-training even for very small training set sizes \cite{rush2012improved}. We also note that syntactically self-trained DCST consistently improves the \biaffine\ parser in our domain adaptation experiments, although the entire training set of the news (nw) section of OntoNotes is used for training.

\begin{figure}[!t]
\centering
\includegraphics[width=1.0\linewidth]{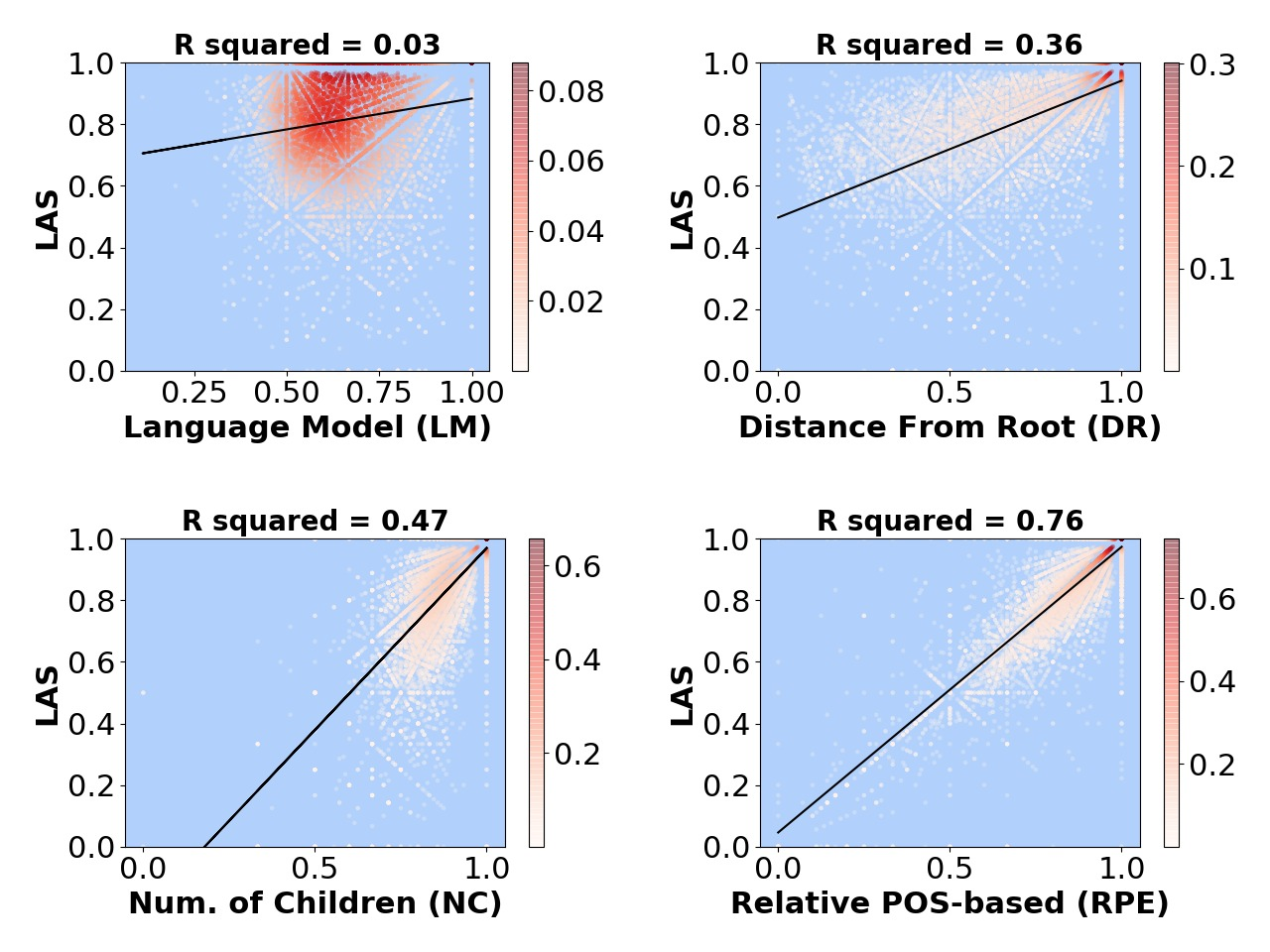}
\caption{Auxiliary task accuracy scores of each BiLSTM tagger vs. the LAS score of the \biaffine\ parser when integrated with that BiLSTM. The BiLSTM scores are computed on the test sets and reflect the capability of the BiLSTM that was trained on unlabeled data with syntactic signal extracted from the base parser's trees (or as a language model for DCST-LM) to properly tag the test sentences. The points correspond to sentence scores across all OntoNotes 5.0 test sets, and the heat map presents the frequency of each point.}
\label{fig:las_vs_accuracy}
\end{figure}

\paragraph{Impact of Self-training Quality}

We next aim to test the connection between the accuracy of the self-trained sequence taggers and the quality of the \biaffine\ parser when integrated with the BiLSTM encoders of these taggers. Ideally, we would expect that the higher the quality of the BiLSTM, the more positive its impact on the parser. This would indicate that the improvement we see with the DCST models indeed results from the information encoded in the self-trained taggers.

To test this hypothesis, Figure~\ref{fig:las_vs_accuracy} plots, for each of the BiLSTM taggers considered in this paper, the sentence-level accuracy scores of the tagger when applied to the OntoNotes test sets vs. the LAS scores of the \biaffine\ parser that was integrated with the corresponding BiLSTM, when that parser was applied to the same test sentences. In such a plot, if the regression line that fits the points has an $R$-squared ($R^2$) value of 1, this indicates a positive linear relation between the self-trained tagger and the parser quality.

The resulting $R^2$ values are well aligned with the relative quality of the DCST models. Particularly, DCST-LM, the least efficient method where the tagger is trained as a language model rather than on a syntactic signal, has an $R^2$ of 0.03. DCST-DR and DCST-NC, which are the next in terms of parsing quality (Table~\ref{table:onto_500}), have $R^2$ values of 0.36 and 0.47, respectively, although DCST-DR performs slightly better. Finally, DCST-RPE, the best performing model among the four in all cases but two, has an $R^2$ value of 0.76. These results provide a positive indication to the hypothesis that the improved parsing quality is caused by the representation model and is not a mere artifact.


\begin{table}[!ht]
\def\arraystretch{0.9}
\scalebox{0.75}{
\centering
\begin{tabular}{c cccc}
\textbf{Model} & AD-NC          & AD-DR          & AD-PDH        & POS Head Error \\\\
\multicolumn{5}{c}{\textbf{OntoNotes}}                                            \\ 
\hline
Base           & 0.305          & 0.539          & 1.371          & 0.162          \\
DCST-NC        & 0.274          & 0.510          & 1.196          & 0.146          \\
DCST-DR        & 0.264          & 0.460          & \textbf{1.099} & 0.141          \\
DCST-RPE       & 0.263          & 0.475          & 1.128          & 0.137          \\
DCST-ENS       & \textbf{0.257} & \textbf{0.458} & 1.121          & \textbf{0.135} \\\\
\multicolumn{5}{c}{\textbf{UD}}                                                    \\
\hline
Base           & 0.366          & 0.600          & 1.377          & 0.163          \\
DCST-NC        & 0.327          & 0.551          & 1.168          & 0.148          \\
DCST-DR        & 0.322          & 0.538          & 1.135          & 0.146         \\
DCST-RPE       & 0.316          & 0.534          & 1.137          & 0.141          \\
DCST-ENS       & \textbf{0.312}          & \textbf{0.524}          & \textbf{1.128}          & \textbf{0.139}         
\end{tabular}}
\caption{Tagging scheme error analysis.}
\label{table:tagging_scheme_error}
\end{table}

\begin{table*}
\def\arraystretch{0.9}
\scalebox{0.75}{
\centering
\begin{tabular}{cllllllllllllll}
&
\multicolumn{2}{c}{bc} &
\multicolumn{2}{c}{bn} &
\multicolumn{2}{c}{mz} &
\multicolumn{2}{c}{nw} &
\multicolumn{2}{c}{pt} &
\multicolumn{2}{c}{tc} &
\multicolumn{2}{c}{wb} \\
\cmidrule(lr){2-3}
\cmidrule(lr){4-5}
\cmidrule(lr){6-7}
\cmidrule(lr){8-9}
\cmidrule(lr){10-11}
\cmidrule(lr){12-13}
\cmidrule(lr){14-15}
{\bf Model} & UAS & LAS & UAS & LAS & UAS & LAS & UAS & LAS & UAS & LAS & UAS & LAS & UAS & LAS \\
\cmidrule(lr){2-3}
\cmidrule(lr){4-5}
\cmidrule(lr){6-7}
\cmidrule(lr){8-9}
\cmidrule(lr){10-11}
\cmidrule(lr){12-13}
\cmidrule(lr){14-15}
Base+ELMo
& 77.96	& 73.97
& 83.12	& 80.18
& 84.62	& 81.37
& 83.09	& 80.35
& 88.82	& 85.55
& 73.84	& 69.23
& 79.67	& 75.77 \\
Base+ELMo+G
& 74.47	& 70.91
& 80.42	& 77.45
& 81.15	& 78.41
& 80.91	& 78.24
& 87.73	& 84.92
& 70.19	& 66.78
& 76.02	& 72.68 \\
DCST-ENS+ELMo
& \underline{\textbf{80.00}}	& \textbf{75.94}
& \underline{\textbf{85.02}}	& \underline{\textbf{81.98}}
& \underline{\textbf{86.24}}	& \textbf{82.54}
& \underline{\textbf{84.56}}	& \underline{\textbf{81.91}}
& \underline{\textbf{90.27}}	& \textbf{86.86}
& \underline{\textbf{77.68}}	& \underline{\textbf{72.72}}
& \underline{\textbf{82.00}}	& \underline{\textbf{77.93}} \\
\end{tabular}}
\caption{Lightly supervised OntoNotes results with ELMo embeddings. \com{The best result for each setup and measure is highlighted in bold.} \com{Underscored results are significant compared to the highest scoring baseline, according to t-test with $p < 0.05$.}}
\label{table:elmo_onto_500}
\end{table*}

\begin{table*}
\def\arraystretch{0.9}
\scalebox{0.56}{
\centering
\begin{tabular}{c llllllllllllllllllll}
{}    &
\multicolumn{2}{c}{cu} &
\multicolumn{2}{c}{da} &
\multicolumn{2}{c}{fa} &
\multicolumn{2}{c}{id} &
\multicolumn{2}{c}{lv} &
\multicolumn{2}{c}{sl} &
\multicolumn{2}{c}{sv} &
\multicolumn{2}{c}{tr} &
\multicolumn{2}{c}{ur} &
\multicolumn{2}{c}{vi} \\
\cmidrule(lr){2-3}
\cmidrule(lr){4-5}
\cmidrule(lr){6-7}
\cmidrule(lr){8-9}
\cmidrule(lr){10-11}
\cmidrule(lr){12-13}
\cmidrule(lr){14-15}
\cmidrule(lr){16-17}
\cmidrule(lr){18-19}
\cmidrule(lr){20-21}

{\bf Model} & UAS & LAS & UAS & LAS & UAS & LAS & UAS & LAS & UAS & LAS & UAS & LAS & UAS & LAS & UAS & LAS & UAS & LAS & UAS & LAS   \\
\cmidrule(lr){2-3}
\cmidrule(lr){4-5}
\cmidrule(lr){6-7}
\cmidrule(lr){8-9}
\cmidrule(lr){10-11}
\cmidrule(lr){12-13}
\cmidrule(lr){14-15}
\cmidrule(lr){16-17}
\cmidrule(lr){18-19}
\cmidrule(lr){20-21}
Base+ELMo
& 72.35	& 61.43
& 80.32	& 76.86
& 85.84	& 81.71
& 73.68	& 58.01
& 79.93	& 73.91
& 76.40	& 67.52
& 81.51	& 76.10
& 53.36	& 34.67
& 86.11	& 79.91
& 71.28	& 67.04 \\
Base+ELMo+G
& \underline{\textbf{75.47}}	& \underline{\textbf{67.07}}
& 79.12	& 75.05
& 83.09	& 79.43
& 73.00	& 57.69
& 72.86	& 67.13
& 74.99	& 69.75
& 79.66	& 74.29
& 53.87	& \underline{\textbf{39.30}}
& 84.83	& 78.53
& 66.57	& 61.56 \\
DCST-ENS+ELMo
& 73.90	& 61.62
& \underline{\textbf{82.29}}	& \textbf{78.49}
& \underline{\textbf{87.87}}	& \textbf{83.25}
& \underline{\textbf{74.95}}	& \underline{\textbf{58.55}}
& \underline{\textbf{82.47}}	& \textbf{76.41}
& \underline{\textbf{79.69}}	& \textbf{70.36}
& \underline{\textbf{83.93}}	& \underline{\textbf{78.27}}
& \underline{\textbf{59.35}}	& 36.81
& \underline{\textbf{87.51}}	& \underline{\textbf{81.53}}
& \underline{\textbf{72.76}}	& \underline{\textbf{68.48}} \\
\end{tabular}}

\caption{Lightly supervised UD results with ELMo embeddings. \com{The best result for each setup and measure is highlighted in bold.} \com{Underscored results are significant compared to the highest scoring baseline, according to t-test with $p < 0.05$.} 
}
\label{table:elmo_ud_500}
\end{table*}

\paragraph{Tagging Scheme Quality Analysis}

We next aim to shed more light on the quality of the tagging schemes with which we train our BiLSTM taggers. We perform an error analysis on the parse trees produced by the final hybrid parser (Figure \ref{fig:sldp_prs_cls_gating}), when each of the schemes is employed in the BiLSTM tagger training step during the lightly-supervised setups. The metrics we compute correspond to the three tagging schemes, and our goal is to examine whether each of the self-trained representation models (BiLSTMs) improves the capability of the final parser to capture the information encoded in its tagging scheme. 

Particularly, we consider four metrics: \textit{Absolute Difference of Number of Children (AD-NC)}: The absolute difference between the number of children a word has in the gold tree and the corresponding number in the predicted tree; \textit{Absolute Difference of Distance from the Root (AD-DR)}: The absolute difference between the distance of a word from the root in the gold tree and the corresponding distance in the predicted tree;
\textit{Absolute Difference of Positional Distance from the Head (AD-PDH)}: The absolute difference between the positional distance of a word from its head word according to the gold tree and the corresponding number according to the predicted tree \cite{kiperwasser-ballesteros-2018-scheduled} (we count the words that separate the head from the modifier in the sentence, considering the distance negative if the word is to the right of its head);
and \textit{POS Head Error}: an indicator function which returns 0 if the POS tag of the head word of a given word according to the gold tree is identical to the corresponding POS tag in the predicted tree, and 1 otherwise.

For all the metrics we report the mean value across all words in our test sets. The values of AD-NC, AD-DR and AD-PDH are hence in the $[0,M]$ range, where $M$ is the length of the longest sentence in the corpus. The values of the POS Head Error are in the $[0,1]$ range. For all metrics lower values indicate that the relevant information has been better captured by the final hybrid parser.

Table \ref{table:tagging_scheme_error} presents a comparison between the Base parser to our DCST algorithms. All in all, the DCST models outperform the Base parser across all comparisons, with DCST-ENS being the best model in all 8 cases except from one. The analysis indicates that in some cases a BiLSTM tagger with a given tagging scheme directly improves the capability of the final parser to capture the corresponding information. For example, DCST-DR, whose tagging scheme considers the distance of each word from the root of the tree, performs best (OntoNotes) or second best (UD) on the AD-DR metric compared to all other models except from the DCST-ENS model that contains the DCST-DR model as a component. Likewise, DCST-RPE, that encodes information about the POS tag of the head word for every word in the sentence,  is the best performing model in terms of POS Head Error. In contrast to the relative success of DCST-RPE and DCST-DR in improving specific capabilities of the parser,  DCST-NC, our weakest model across experimental setups, is also the weakest DCST model in this error analysis, even when considering the AD-NC metric that measures success in predicting the number of children a word has in the tree.


\paragraph{Sentence Length Adaptation}

\begin{table}
\small
\centering
\begin{tabular}{ccc}
{\bf{Model}} &  UAS & LAS               \\
\hline
Base                      & 54.86 & 52.65                              \\
DCST-LM                   & 55.26 & 52.63                              \\
Self-Training             & 54.22 & 52.16
                        \\
CVT                       & 50.61 & 46.13                              \\
\midrule

DCST-ENS                  & \underline{\textbf{58.85}} & \underline{\textbf{56.64}} \\                        
\end{tabular}
\caption{Sentence length adaptation results. \com{The best result for each setup and measure is highlighted in bold.}\com{: Models are trained on Ontonotes web (wb) sentences with up to 10 words and are tested on web (wb) sentences with more than 10 words.} \com{Underscored results are significant compared to the baseline with the highest score, according to t-test with $p < 0.05$}}
\label{table:wb_under_over_10}
\end{table}

We next aim to test whether DCST can enhance a parser trained on short sentences so that it can better parse long sentences. Dependency parsers perform better on short sentences, and we would expect self-training to bring in high quality syntactic information from automatically parsed long sentences.

For this aim, we replicate the Onotnotes wb in-domain experiment, except that we train the parser on all training set sentences of up to 10 words, use the training set sentences with more than 10 words as unlabeled data for sequence tagger training (Algorithm 1, step 4), and test the final parser on all test sentences with more than 10 words. 

Table~\ref{table:wb_under_over_10} shows that DCST-ENS improves the Base parser in this setup by 3.99 UAS and LAS points. DCST-LM achieves only a marginal UAS improvement while CVT substantially harms the parser. This result further supports the value of our methods and encourages future research in various under-resourced setups.

\paragraph{ELMo Embeddings}

Finally, we turn to investigate the impact of deep contextualized word embeddings, such as ELMo \cite{peters-etal-2018-deep}, on the base parser and on the DCST-ENS model. To this end, we replace the Glove/FastText word embeddings from our original experiments with the multilingual ELMo word embeddings of \citet{che2018towards}. We follow \citet{che2018towards} and define the ELMo word embedding for word $i$ as: $w_i = W^{ELMo}\cdot\frac{1}{3}\sum_{j=0}^{2}h_{i,j}^{ELMo}$, where $W^{ELMo}$ is a trainable parameter and $h_{i,j}^{ELMo}$ is the hidden representation for word $i$ in the $j$'th BiLSTM layer of the ELMo model, which remains fixed throughout all experiments. 

We experiment with three models: \textbf{Base + ELMo}: the \biaffine\ parser fed by the ELMo word embeddings and trained on the labeled training data; \textbf{Base + ELMo + Gating (G)}: The \biaffine\ parser fed by our original word embeddings, and ELMo word embeddings are integrated through our gating mechanism. Training is done on the labeled training data only; and \textbf{DCST-ENS + ELMo}: our ensemble parser where the BiLSTM taggers and the Base parser are fed by the ELMo word embeddings. 

Tables \ref{table:elmo_onto_500} (OntoNotes) and \ref{table:elmo_ud_500} (UD) summarize the results in the lightly supervised setups with 500 training sentences. Like in previous experiments, DCST-ENS+ELMo is the best performing model in both setups. While Base+ELMo+G is superior in the $cu$ and $tr$ (LAS) setups, it is inferior in all OntoNotes domains. Note also that DCST-ENS+ELMo improves the UAS results of DCST-ENS from tables \ref{table:onto_500} and \ref{table:ud_500} on all OntoNotes domains and on 7 out of 10 UD languages. 

\section{Conclusions}
\label{sec:discussion}

We proposed a new self-training framework for dependency parsing. Our DCST approach is based on the integration of (a) contextualized embedding model(s) into a neural dependency parser, where the embedding models are trained on word tagging schemes extracted from the trees generated by the base parser on unlabeled data. In multilingual lightly-supervised and domain adaptation experiments, our models consistently outperform strong baselines and previous models.


In future work we intend to explore improved word tagging schemes, sequence tagging architectures and integration mechanisms. We shall also consider cross-language learning where the lexical gap between languages should be overcome.


\section*{Acknowledgements}

We would like to thank the action editor and the reviewers, as well as the members of the IE@Technion NLP group for their valuable feedback and advice. This research was partially funded by an ISF personal grant No. 1625/18.

\bibliography{tacl2018}

\begin{thebibliography}{61}
\expandafter\ifx\csname natexlab\endcsname\relax\def\natexlab#1{#1}\fi

\bibitem[{Abney(2004)}]{abney2004understanding}
Steven Abney. 2004.
\newblock \href {https://www.aclweb.org/anthology/J04-3004} {Understanding the
  {Yarowsky} algorithm}.
\newblock \emph{Computational Linguistics}, 30(3):365--395.

\bibitem[{Angeli et~al.(2015)Angeli, Premkumar, and Manning}]{Angeli15}
Gabor Angeli, Melvin Jose~Johnson Premkumar, and Christopher~D. Manning. 2015.
\newblock \href {http://aclweb.org/anthology/P/P15/P15-1034.pdf} {Leveraging
  linguistic structure for open domain information extraction}.
\newblock In \emph{Proceedings of {ACL}}.

\bibitem[{Artetxe et~al.(2018)Artetxe, Labaka, and
  Agirre}]{artetxe-etal-2018-robust}
Mikel Artetxe, Gorka Labaka, and Eneko Agirre. 2018.
\newblock \href {https://www.aclweb.org/anthology/P18-1073} {A robust
  self-learning method for fully unsupervised cross-lingual mappings of word
  embeddings}.
\newblock In \emph{Proceedings of the 56th Annual Meeting of the Association
  for Computational Linguistics (Volume 1: Long Papers)}, pages 789--798,
  Melbourne, Australia. Association for Computational Linguistics.

\bibitem[{Blum and Mitchell(1998)}]{Blum:98}
Avrim Blum and Tom Mitchell. 1998.
\newblock \href
  {http://www1.cs.columbia.edu/~dplewis/candidacy/blum98combining.pdf}
  {Combining labeled and unlabeled data with co-training}.
\newblock In \emph{Proceedings of the eleventh annual conference on
  Computational learning theory}, pages 92--100. ACM.

\bibitem[{Che et~al.(2018)Che, Liu, Wang, Zheng, and Liu}]{che2018towards}
Wanxiang Che, Yijia Liu, Yuxuan Wang, Bo~Zheng, and Ting Liu. 2018.
\newblock \href {https://www.aclweb.org/anthology/K18-2005} {Towards better ud
  parsing: Deep contextualized word embeddings, ensemble, and treebank
  concatenation}.
\newblock In \emph{Proceedings of the CoNLL 2018 Shared Task: Multilingual
  Parsing from Raw Text to Universal Dependencies}, pages 55--64.

\bibitem[{Chen et~al.(2008)Chen, Wu, and Isahara}]{chen2008learning}
Wenliang Chen, Youzheng Wu, and Hitoshi Isahara. 2008.
\newblock \href {https://www.aclweb.org/anthology/C08-1015} {Learning reliable
  information for dependency parsing adaptation}.
\newblock In \emph{Proceedings of the 22nd International Conference on
  Computational Linguistics-Volume 1}, pages 113--120. Association for
  Computational Linguistics.

\bibitem[{Chen et~al.(2014)Chen, Zhang, and Zhang}]{chen2014feature}
Wenliang Chen, Yue Zhang, and Min Zhang. 2014.
\newblock \href {https://www.aclweb.org/anthology/C14-1078} {Feature embedding
  for dependency parsing}.
\newblock In \emph{Proceedings of COLING 2014, the 25th International
  Conference on Computational Linguistics: Technical Papers}, pages 816--826.

\bibitem[{Clark et~al.(2018)Clark, Luong, Manning, and Le}]{clark2018semi}
Kevin Clark, Minh-Thang Luong, Christopher~D. Manning, and Quoc~V. Le. 2018.
\newblock \href {https://aclweb.org/anthology/D18-1217} {Semi-supervised
  sequence modeling with cross-view training}.
\newblock In \emph{Proceedings of the 2018 Conference on Empirical Methods in
  Natural Language Processing}, pages 1914--1925.

\bibitem[{Clevert et~al.(2016)Clevert, Unterthiner, and
  Hochreiter}]{clevert2015fast}
Djork{-}Arn{\'{e}} Clevert, Thomas Unterthiner, and Sepp Hochreiter. 2016.
\newblock \href {http://arxiv.org/abs/1511.07289} {Fast and accurate deep
  network learning by exponential linear units ({ELU}s)}.
\newblock In \emph{4th International Conference on Learning Representations,
  {ICLR} 2016, San Juan, Puerto Rico, May 2-4, 2016, Conference Track
  Proceedings}.

\bibitem[{Devlin et~al.(2019)Devlin, Chang, Lee, and
  Toutanova}]{devlin2019bert}
Jacob Devlin, Ming-Wei Chang, Kenton Lee, and Kristina Toutanova. 2019.
\newblock \href {https://doi.org/10.18653/v1/N19-1423} {{BERT}: Pre-training of
  deep bidirectional transformers for language understanding}.
\newblock In \emph{Proceedings of the 2019 Conference of the North {A}merican
  Chapter of the Association for Computational Linguistics: Human Language
  Technologies, Volume 1 (Long and Short Papers)}, pages 4171--4186,
  Minneapolis, Minnesota. Association for Computational Linguistics.

\bibitem[{Dozat and Manning(2017)}]{Dozat:17}
Timothy Dozat and Christopher~D. Manning. 2017.
\newblock \href
  {https://web.stanford.edu/~tdozat/files/TDozat-ICLR2017-Paper.pdf} {Deep
  biaffine attention for neural dependency parsing}.
\newblock In \emph{5th International Conference on Learning Representations,
  {ICLR} 2017, Toulon, France, April 24-26, 2017, Conference Track
  Proceedings}.

\bibitem[{Dror et~al.(2018)Dror, Baumer, Shlomov, and
  Reichart}]{dror-etal-2018-hitchhikers}
Rotem Dror, Gili Baumer, Segev Shlomov, and Roi Reichart. 2018.
\newblock \href {https://www.aclweb.org/anthology/P18-1128} {The hitchhiker{'}s
  guide to testing statistical significance in natural language processing}.
\newblock In \emph{Proceedings of the 56th Annual Meeting of the Association
  for Computational Linguistics (Volume 1: Long Papers)}, pages 1383--1392,
  Melbourne, Australia. Association for Computational Linguistics.

\bibitem[{Edmonds(1967)}]{edmonds1967optimum}
Jack Edmonds. 1967.
\newblock \href
  {https://nvlpubs.nist.gov/nistpubs/jres/71b/jresv71bn4p233_a1b.pdf} {Optimum
  branchings}.
\newblock \emph{Journal of Research of the national Bureau of Standards B},
  71(4):233--240.

\bibitem[{Goldwasser et~al.(2011)Goldwasser, Reichart, Clarke, and
  Roth}]{Goldwasser:11}
Dan Goldwasser, Roi Reichart, James Clarke, and Dan Roth. 2011.
\newblock \href {https://www.aclweb.org/anthology/P11-1149} {Confidence driven
  unsupervised semantic parsing}.
\newblock In \emph{Proceedings of the 49th Annual Meeting of the Association
  for Computational Linguistics: Human Language Technologies}, pages
  1486--1495.

\bibitem[{Grave et~al.(2018)Grave, Bojanowski, Gupta, Joulin, and
  Mikolov}]{grave2018learning}
Edouard Grave, Piotr Bojanowski, Prakhar Gupta, Armand Joulin, and Tomas
  Mikolov. 2018.
\newblock \href {https://www.aclweb.org/anthology/L18-1550} {Learning word
  vectors for 157 languages}.
\newblock In \emph{Proceedings of the International Conference on Language
  Resources and Evaluation (LREC 2018)}.

\bibitem[{Hadiwinoto and Ng(2017)}]{Hadiwinoto17}
Christian Hadiwinoto and Hwee~Tou Ng. 2017.
\newblock \href {http://aaai.org/ocs/index.php/AAAI/AAAI17/paper/view/14878} {A
  dependency-based neural reordering model for statistical machine
  translation}.
\newblock In \emph{Thirty-First AAAI Conference on Artificial Intelligence}.

\bibitem[{He and Zhou(2011)}]{he2011self}
Yulan He and Deyu Zhou. 2011.
\newblock \href
  {https://research.aston.ac.uk/portal/files/3337561/Self_training_from_labeled_features_for_sentiment_analysis_2011.pdf}
  {Self-training from labeled features for sentiment analysis}.
\newblock \emph{Information Processing \& Management}, 47(4):606--616.

\bibitem[{Hershcovich et~al.(2017)Hershcovich, Abend, and
  Rappoport}]{Hershcovich:17}
Daniel Hershcovich, Omri Abend, and Ari Rappoport. 2017.
\newblock \href {https://www.aclweb.org/anthology/P17-1104} {A transition-based
  directed acyclic graph parser for {UCCA}}.
\newblock In \emph{Proceedings of the 55th Annual Meeting of the Association
  for Computational Linguistics (Volume 1: Long Papers)}, volume~1, pages
  1127--1138.

\bibitem[{Hovy et~al.(2006)Hovy, Marcus, Palmer, Ramshaw, and
  Weischedel}]{hovy2006ontonotes}
Eduard Hovy, Mitchell Marcus, Martha Palmer, Lance Ramshaw, and Ralph
  Weischedel. 2006.
\newblock \href {https://www.aclweb.org/anthology/N06-2015} {Ontonotes: The
  90\% solution}.
\newblock In \emph{Proceedings of the human language technology conference of
  the NAACL, Companion Volume: Short Papers}.

\bibitem[{Imamura and Sumita(2018)}]{imamura2018nict}
Kenji Imamura and Eiichiro Sumita. 2018.
\newblock \href {https://www.aclweb.org/anthology/W18-2713} {{NICT}
  self-training approach to neural machine translation at nmt-2018}.
\newblock In \emph{Proceedings of the 2nd Workshop on Neural Machine
  Translation and Generation}, pages 110--115.

\bibitem[{Kingma and Ba(2015)}]{kingma2014adam}
Diederik~P. Kingma and Jimmy Ba. 2015.
\newblock \href {https://arxiv.org/abs/1412.6980} {Adam: A method for
  stochastic optimization}.
\newblock In \emph{Proceedings of ICLR}.

\bibitem[{Kiperwasser and
  Ballesteros(2018)}]{kiperwasser-ballesteros-2018-scheduled}
Eliyahu Kiperwasser and Miguel Ballesteros. 2018.
\newblock \href {https://doi.org/10.1162/tacl_a_00017} {Scheduled multi-task
  learning: From syntax to translation}.
\newblock \emph{Transactions of the Association for Computational Linguistics},
  6:225--240.

\bibitem[{Kiperwasser and Goldberg(2016)}]{Kiperwasser16}
Eliyahu Kiperwasser and Yoav Goldberg. 2016.
\newblock \href {https://transacl.org/ojs/index.php/tacl/article/view/885}
  {Simple and accurate dependency parsing using bidirectional {LSTM} feature
  representations}.
\newblock \emph{Transactions of the Association for Computational Linguistics},
  4:313--327.

\bibitem[{Levy and Goldberg(2014)}]{levy2014dependency}
Omer Levy and Yoav Goldberg. 2014.
\newblock \href {http://aclweb.org/anthology/P14-2050} {Dependency-based word
  embeddings}.
\newblock In \emph{Proceedings of the 52nd Annual Meeting of the Association
  for Computational Linguistics (Volume 2: Short Papers)}, volume~2, pages
  302--308.

\bibitem[{Ma et~al.(2018)Ma, Hu, Liu, Peng, Neubig, and Hovy}]{ma2018stack}
Xuezhe Ma, Zecong Hu, Jingzhou Liu, Nanyun Peng, Graham Neubig, and Eduard
  Hovy. 2018.
\newblock \href {https://www.aclweb.org/anthology/P18-1130} {Stack-pointer
  networks for dependency parsing}.
\newblock In \emph{Proceedings of the 56th Annual Meeting of the Association
  for Computational Linguistics (Volume 1: Long Papers)}, pages 1403--1414.

\bibitem[{Marcheggiani et~al.(2017)Marcheggiani, Frolov, and
  Titov}]{Marcheggiani17}
Diego Marcheggiani, Anton Frolov, and Ivan Titov. 2017.
\newblock \href {https://doi.org/10.18653/v1/K17-1041} {A simple and accurate
  syntax-agnostic neural model for dependency-based semantic role labeling}.
\newblock In \emph{Proceedings of {CoNLL}}.

\bibitem[{McCann et~al.(2017)McCann, Bradbury, Xiong, and
  Socher}]{mccann2017learned}
Bryan McCann, James Bradbury, Caiming Xiong, and Richard Socher. 2017.
\newblock \href
  {https://papers.nips.cc/paper/7209-learned-in-translation-contextualized-word-vectors.pdf}
  {Learned in translation: Contextualized word vectors}.
\newblock In \emph{Advances in Neural Information Processing Systems}, pages
  6294--6305.

\bibitem[{McClosky et~al.(2006{\natexlab{a}})McClosky, Charniak, and
  Johnson}]{mcclosky2006effective}
David McClosky, Eugene Charniak, and Mark Johnson. 2006{\natexlab{a}}.
\newblock \href {https://aclweb.org/anthology/N06-1020} {Effective
  self-training for parsing}.
\newblock In \emph{Proceedings of the main conference on human language
  technology conference of the North American Chapter of the Association of
  Computational Linguistics}, pages 152--159. Association for Computational
  Linguistics.

\bibitem[{McClosky et~al.(2006{\natexlab{b}})McClosky, Charniak, and
  Johnson}]{mcclosky2006reranking}
David McClosky, Eugene Charniak, and Mark Johnson. 2006{\natexlab{b}}.
\newblock \href {https://aclweb.org/anthology/P06-1043} {Reranking and
  self-training for parser adaptation}.
\newblock In \emph{Proceedings of the 21st International Conference on
  Computational Linguistics and the 44th annual meeting of the Association for
  Computational Linguistics}, pages 337--344. Association for Computational
  Linguistics.

\bibitem[{McClosky et~al.(2010)McClosky, Charniak, and
  Johnson}]{mcclosky2010automatic}
David McClosky, Eugene Charniak, and Mark Johnson. 2010.
\newblock \href {https://aclweb.org/anthology/N10-1004} {Automatic domain
  adaptation for parsing}.
\newblock In \emph{Human Language Technologies: The 2010 Annual Conference of
  the North American Chapter of the Association for Computational Linguistics},
  pages 28--36. Association for Computational Linguistics.

\bibitem[{McDonald et~al.(2013)McDonald, Nivre, Quirmbach-Brundage, Goldberg,
  Das, Ganchev, Hall, Petrov, Zhang, T{\"a}ckstr{\"o}m, Bedini, Castell{\'o},
  and Lee}]{mcdonald2013universal}
Ryan McDonald, Joakim Nivre, Yvonne Quirmbach-Brundage, Yoav Goldberg, Dipanjan
  Das, Kuzman Ganchev, Keith Hall, Slav Petrov, Hao Zhang, Oscar
  T{\"a}ckstr{\"o}m, Claudia Bedini, N{\'u}ria~Bertomeu Castell{\'o}, and
  Jungmee Lee. 2013.
\newblock \href {https://www.aclweb.org/anthology/P13-2017} {Universal
  dependency annotation for multilingual parsing}.
\newblock In \emph{Proceedings of the 51st Annual Meeting of the Association
  for Computational Linguistics (Volume 2: Short Papers)}, volume~2, pages
  92--97.

\bibitem[{Mihalcea(2004)}]{mihalcea2004co}
Rada Mihalcea. 2004.
\newblock \href {https://www.aclweb.org/anthology/W04-2405} {Co-training and
  self-training for word sense disambiguation}.
\newblock In \emph{Proceedings of the Eighth Conference on Computational
  Natural Language Learning (CoNLL-2004) at HLT-NAACL 2004}.

\bibitem[{Nivre et~al.(2018)Nivre, Abrams, Agi{\'c}, Ahrenberg, Antonsen,
  Aranzabe, Arutie, Asahara, Ateyah, Attia et~al.}]{nivre2018universal}
Joakim Nivre, Mitchell Abrams, {\v{Z}}eljko Agi{\'c}, Lars Ahrenberg, Lene
  Antonsen, Maria~Jesus Aranzabe, Gashaw Arutie, Masayuki Asahara, Luma Ateyah,
  Mohammed Attia, et~al. 2018.
\newblock \href
  {https://hal-univ-tlse3.archives-ouvertes.fr/CERHAC/hal-01930733v1}
  {Universal dependencies 2.2}.

\bibitem[{Nivre et~al.(2016)Nivre, De~Marneffe, Ginter, Goldberg, Hajic,
  Manning, McDonald, Petrov, Pyysalo, Silveira, Tsarfaty, and
  Zeman}]{nivre2016universal}
Joakim Nivre, Marie-Catherine De~Marneffe, Filip Ginter, Yoav Goldberg, Jan
  Hajic, Christopher~D. Manning, Ryan McDonald, Slav Petrov, Sampo Pyysalo,
  Natalia Silveira, Reut Tsarfaty, and Daniel Zeman. 2016.
\newblock \href {https://nlp.stanford.edu/pubs/nivre2016ud.pdf} {Universal
  dependencies v1: A multilingual treebank collection.}
\newblock In \emph{LREC}.

\bibitem[{Pennington et~al.(2014)Pennington, Socher, and
  Manning}]{pennington2014glove}
Jeffrey Pennington, Richard Socher, and Christopher~D. Manning. 2014.
\newblock \href {https://www.aclweb.org/anthology/D14-1162} {Glo{V}e: Global
  vectors for word representation}.
\newblock In \emph{Proceedings of the 2014 conference on empirical methods in
  natural language processing (EMNLP)}, pages 1532--1543.

\bibitem[{Peters et~al.(2018)Peters, Neumann, Iyyer, Gardner, Clark, Lee, and
  Zettlemoyer}]{peters-etal-2018-deep}
Matthew Peters, Mark Neumann, Mohit Iyyer, Matt Gardner, Christopher Clark,
  Kenton Lee, and Luke Zettlemoyer. 2018.
\newblock \href {https://doi.org/10.18653/v1/N18-1202} {Deep contextualized
  word representations}.
\newblock In \emph{Proceedings of the 2018 Conference of the North {A}merican
  Chapter of the Association for Computational Linguistics: Human Language
  Technologies, Volume 1 (Long Papers)}, pages 2227--2237, New Orleans,
  Louisiana. Association for Computational Linguistics.

\bibitem[{Plank and Agi{\'c}(2018)}]{plank-agic-2018-distant}
Barbara Plank and {\v{Z}}eljko Agi{\'c}. 2018.
\newblock \href {https://doi.org/10.18653/v1/D18-1061} {Distant supervision
  from disparate sources for low-resource part-of-speech tagging}.
\newblock In \emph{Proceedings of the 2018 Conference on Empirical Methods in
  Natural Language Processing}, pages 614--620, Brussels, Belgium. Association
  for Computational Linguistics.

\bibitem[{Plank and Van~Noord(2011)}]{plank2011effective}
Barbara Plank and Gertjan Van~Noord. 2011.
\newblock \href {https://www.aclweb.org/anthology/P11-1157} {Effective measures
  of domain similarity for parsing}.
\newblock In \emph{Proceedings of the 49th Annual Meeting of the Association
  for Computational Linguistics: Human Language Technologies-Volume 1}, pages
  1566--1576. Association for Computational Linguistics.

\bibitem[{Reichart and Rappoport(2007)}]{reichart2007self}
Roi Reichart and Ari Rappoport. 2007.
\newblock \href {https://www.aclweb.org/anthology/P07-1078} {Self-training for
  enhancement and domain adaptation of statistical parsers trained on small
  datasets}.
\newblock In \emph{Proceedings of the 45th Annual Meeting of the Association of
  Computational Linguistics}, pages 616--623.

\bibitem[{Ruder and Plank(2018)}]{ruder2018strong}
Sebastian Ruder and Barbara Plank. 2018.
\newblock \href {https://www.aclweb.org/anthology/P18-1096} {Strong baselines
  for neural semi-supervised learning under domain shift}.
\newblock In \emph{The 56th Annual Meeting of the Association for Computational
  LinguisticsMeeting of the Association for Computational Linguistics}.
  Association for Computational Linguistics.

\bibitem[{Rush et~al.(2012)Rush, Reichart, Collins, and
  Globerson}]{rush2012improved}
Alexander~M. Rush, Roi Reichart, Michael Collins, and Amir Globerson. 2012.
\newblock \href {https://www.aclweb.org/anthology/D12-1131} {Improved parsing
  and {POS} tagging using inter-sentence consistency constraints}.
\newblock In \emph{Proceedings of the 2012 Joint Conference on Empirical
  Methods in Natural Language Processing and Computational Natural Language
  Learning}, pages 1434--1444. Association for Computational Linguistics.

\bibitem[{Rybak and Wr{\'o}blewska(2018)}]{rybak2018semi}
Piotr Rybak and Alina Wr{\'o}blewska. 2018.
\newblock \href
  {https://universaldependencies.org/conll18/proceedings/pdf/K18-2004.pdf}
  {Semi-supervised neural system for tagging, parsing and lematization}.
\newblock In \emph{Proceedings of the CoNLL 2018 Shared Task: Multilingual
  Parsing from Raw Text to Universal Dependencies}, pages 45--54.

\bibitem[{Sato et~al.(2017)Sato, Manabe, Noji, and
  Matsumoto}]{sato2017adversarial}
Motoki Sato, Hitoshi Manabe, Hiroshi Noji, and Yuji Matsumoto. 2017.
\newblock \href {https://www.aclweb.org/anthology/K17-3007} {Adversarial
  training for cross-domain universal dependency parsing}.
\newblock In \emph{Proceedings of the CoNLL 2017 Shared Task: Multilingual
  Parsing from Raw Text to Universal Dependencies}, pages 71--79.

\bibitem[{Shareghi et~al.(2019)Shareghi, Li, Zhu, Reichart, and
  Korhonen}]{shareghi2019bayesian}
Ehsan Shareghi, Yingzhen Li, Yi~Zhu, Roi Reichart, and Anna Korhonen. 2019.
\newblock \href {https://www.aclweb.org/anthology/N19-1354} {Bayesian learning
  for neural dependency parsing}.
\newblock In \emph{Proceedings of the 2019 Conference of the North American
  Chapter of the Association for Computational Linguistics: Human Language
  Technologies, Volume 1 (Long and Short Papers)}, pages 3509--3519.

\bibitem[{S{\o}gaard(2010)}]{Sogaard:10}
Anders S{\o}gaard. 2010.
\newblock \href {https://www.aclweb.org/anthology/P10-2038} {Simple
  semi-supervised training of part-of-speech taggers}.
\newblock In \emph{Proceedings of the ACL 2010 Conference Short Papers}, pages
  205--208. Association for Computational Linguistics.

\bibitem[{Spoustov{\'a} and Spousta(2010)}]{spoustova2010dependency}
Drahom{\'\i}ra Spoustov{\'a} and Miroslav Spousta. 2010.
\newblock \href
  {https://www.degruyter.com/downloadpdf/j/pralin.2010.94.issue--1/v10108-010-0017-3/v10108-010-0017-3.pdf}
  {Dependency parsing as a sequence labeling task}.
\newblock \emph{The Prague Bulletin of Mathematical Linguistics}, 94:7--14.

\bibitem[{Srivastava et~al.(2014)Srivastava, Hinton, Krizhevsky, Sutskever, and
  Salakhutdinov}]{srivastava2014dropout}
Nitish Srivastava, Geoffrey Hinton, Alex Krizhevsky, Ilya Sutskever, and Ruslan
  Salakhutdinov. 2014.
\newblock \href
  {http://www.jmlr.org/papers/volume15/srivastava14a/srivastava14a.pdf?utm_content=buffer79b43&utm_medium=social&utm_source=twitter.com&utm_campaign=buffer}
  {Dropout: a simple way to prevent neural networks from overfitting}.
\newblock \emph{The Journal of Machine Learning Research}, 15(1):1929--1958.

\bibitem[{Steedman et~al.(2003)Steedman, Osborne, Sarkar, Clark, Hwa,
  Hockenmaier, Ruhlen, Baker, and Crim}]{steedman2003bootstrapping}
Mark Steedman, Miles Osborne, Anoop Sarkar, Stephen Clark, Rebecca Hwa, Julia
  Hockenmaier, Paul Ruhlen, Steven Baker, and Jeremiah Crim. 2003.
\newblock \href {https://aclweb.org/anthology/E03-1008} {Bootstrapping
  statistical parsers from small datasets}.
\newblock In \emph{Proceedings of the tenth conference on European chapter of
  the Association for Computational Linguistics-Volume 1}, pages 331--338.
  Association for Computational Linguistics.

\bibitem[{Strzyz et~al.(2019)Strzyz, Vilares, and
  G{\'o}mez-Rodr{\i}guez}]{strzyz2019viable}
Michalina Strzyz, David Vilares, and Carlos G{\'o}mez-Rodr{\i}guez. 2019.
\newblock \href {https://www.aclweb.org/anthology/N19-1077} {Viable dependency
  parsing as sequence labeling}.
\newblock In \emph{Proceedings of NAACL-HLT}, pages 717--723.

\bibitem[{Tenney et~al.(2019)Tenney, Xia, Chen, Wang, Poliak, McCoy, Kim,
  Van~Durme, Bowman, Das, and Pavlick}]{Tenney:2019}
Ian Tenney, Patrick Xia, Berlin Chen, Alex Wang, Adam Poliak, R.~Thomas McCoy,
  Najoung Kim, Benjamin Van~Durme, Samuel~R. Bowman, Dipanjan Das, and Ellie
  Pavlick. 2019.
\newblock \href {https://arxiv.org/abs/1905.06316} {What do you learn from
  context? probing for sentence structure in contextualized word
  representations}.
\newblock In \emph{Proceedings of ICLR}.

\bibitem[{Toutanova et~al.(2016)Toutanova, Lin, Yih, Poon, and
  Quirk}]{Toutanova16}
Kristina Toutanova, Xi~Victoria Lin, Wen-tau Yih, Hoifung Poon, and Chris
  Quirk. 2016.
\newblock \href {http://aclweb.org/anthology/P/P16/P16-1136.pdf} {Compositional
  learning of embeddings for relation paths in knowledge base and text}.
\newblock In \emph{Proceedings of the 54th Annual Meeting of the Association
  for Computational Linguistics (Volume 1: Long Papers)}, volume~1, pages
  1434--1444.

\bibitem[{Vaswani et~al.(2017)Vaswani, Shazeer, Parmar, Uszkoreit, Jones,
  Gomez, Kaiser, and Polosukhin}]{vaswani:17}
Ashish Vaswani, Noam Shazeer, Niki Parmar, Jakob Uszkoreit, Llion Jones,
  Aidan~N. Gomez, {\L}ukasz Kaiser, and Illia Polosukhin. 2017.
\newblock \href
  {https://papers.nips.cc/paper/7181-attention-is-all-you-need.pdf} {Attention
  is all you need}.
\newblock In \emph{Advances in neural information processing systems}, pages
  5998--6008.

\bibitem[{Vinyals et~al.(2015)Vinyals, Kaiser, Koo, Petrov, Sutskever, and
  Hinton}]{Vinyals:15}
Oriol Vinyals, {\L}ukasz Kaiser, Terry Koo, Slav Petrov, Ilya Sutskever, and
  Geoffrey Hinton. 2015.
\newblock \href
  {https://papers.nips.cc/paper/5635-grammar-as-a-foreign-language.pdf}
  {Grammar as a foreign language}.
\newblock In \emph{Advances in neural information processing systems}, pages
  2773--2781.

\bibitem[{Wang et~al.(2019)Wang, Hula, Xia, Pappagari, McCoy, Patel, Kim,
  Tenney, Huang, Yu, Jin, Chen, Van~Durme, Grave, Pavlick, and
  Bowman}]{wang2019can}
Alex Wang, Jan Hula, Patrick Xia, Raghavendra Pappagari, R.~Thomas McCoy, Roma
  Patel, Najoung Kim, Ian Tenney, Yinghui Huang, Katherin Yu, Shuning Jin,
  Berlin Chen, Benjamin Van~Durme, Edouard Grave, Ellie Pavlick, and Samuel~R.
  Bowman. 2019.
\newblock \href {https://www.aclweb.org/anthology/P19-1439} {Can you tell me
  how to get past sesame street? sentence-level pretraining beyond language
  modeling}.
\newblock In \emph{Proceedings of the 57th Conference of the Association for
  Computational Linguistics}, pages 4465--4476.

\bibitem[{Wieting and Kiela(2019)}]{Wieting:19}
John Wieting and Douwe Kiela. 2019.
\newblock \href {https://arxiv.org/abs/1901.10444} {No training required:
  Exploring random encoders for sentence classification}.
\newblock In \emph{Proceedings of ICLR}.

\bibitem[{Yadav and Bethard(2018)}]{Yadav:18}
Vikas Yadav and Steven Bethard. 2018.
\newblock \href {https://www.aclweb.org/anthology/C18-1182} {A survey on recent
  advances in named entity recognition from deep learning models}.
\newblock In \emph{Proceedings of the 27th International Conference on
  Computational Linguistics}, pages 2145--2158.

\bibitem[{Yarowsky(1995)}]{yarowsky1995unsupervised}
David Yarowsky. 1995.
\newblock \href {https://www.aclweb.org/anthology/P95-1026} {Unsupervised word
  sense disambiguation rivaling supervised methods}.
\newblock In \emph{33rd annual meeting of the association for computational
  linguistics}.

\bibitem[{Zhang and Bowman(2018)}]{zhang-bowman-2018-language}
Kelly~W. Zhang and Samuel~R. Bowman. 2018.
\newblock \href {https://doi.org/10.18653/v1/W18-5448} {Language modeling
  teaches you more than translation does: Lessons learned through auxiliary
  syntactic task analysis}.
\newblock In \emph{Proceedings of the 2018 {EMNLP} Workshop {B}lackbox{NLP}:
  Analyzing and Interpreting Neural Networks for {NLP}}, pages 359--361,
  Brussels, Belgium. Association for Computational Linguistics.

\bibitem[{Zhang et~al.(2015)Zhang, Zhao, and LeCun}]{zhang2015character}
Xiang Zhang, Junbo Zhao, and Yann LeCun. 2015.
\newblock \href
  {http://papers.nips.cc/paper/5782-character-level-convolutional-networks-for-text-classification.pdf}
  {Character-level convolutional networks for text classification}.
\newblock In \emph{Advances in neural information processing systems}, pages
  649--657.

\bibitem[{Zhou and Li(2005)}]{Zhou:05}
Zhi-Hua Zhou and Ming Li. 2005.
\newblock \href
  {https://cs.nju.edu.cn/zhouzh/zhouzh.files/publication/tkde05.pdf}
  {Tri-training: Exploiting unlabeled data using three classifiers}.
\newblock \emph{IEEE Transactions on Knowledge \& Data Engineering},
  (11):1529--1541.

\bibitem[{Ziser and Reichart(2018)}]{ziser-reichart-2018-pivot}
Yftah Ziser and Roi Reichart. 2018.
\newblock \href {https://doi.org/10.18653/v1/N18-1112} {Pivot based language
  modeling for improved neural domain adaptation}.
\newblock In \emph{Proceedings of the 2018 Conference of the North {A}merican
  Chapter of the Association for Computational Linguistics: Human Language
  Technologies, Volume 1 (Long Papers)}, pages 1241--1251, New Orleans,
  Louisiana. Association for Computational Linguistics.

\end{thebibliography}
\bibliographystyle{acl_natbib}

\end{document}